\documentclass[sigconf] {acmart}
\usepackage[utf8]{inputenc}
\usepackage{threeparttable}
\usepackage{booktabs}
\usepackage[ruled,linesnumbered]{algorithm2e}
\usepackage{extarrows}
\usepackage{multirow} 
\usepackage{enumitem}

\usepackage{soul}
\usepackage[normalem]{ulem}
\AtBeginDocument{%
  }

\copyrightyear{2026}
\acmYear{2026}
\setcopyright{cc}
\setcctype{by}
\acmConference[KDD '26]{Proceedings of the 32nd ACM SIGKDD Conference on Knowledge Discovery and Data Mining V.1}{August 09--13, 2026}{Jeju Island, Republic of Korea}
\acmBooktitle{Proceedings of the 32nd ACM SIGKDD Conference on Knowledge Discovery and Data Mining V.1 (KDD '26), August 09--13, 2026, Jeju Island, Republic of Korea}
\acmPrice{}
\acmDOI{10.1145/3770854.3780282}
\acmISBN{979-8-4007-2258-5/2026/08}

\theoremstyle{plain} 
\newtheorem{definition}{Definition}
\newtheorem{theorem}{Theorem}
\newtheorem{lemma}{Lemma}
\newtheorem{remark}{Remark}

\begin{document}
\title{Adversarial Signed Graph Learning with Differential Privacy}

\author{Haobin Ke}
\affiliation{%
 \institution{The Hong Kong Polytechnic University}
 \city{Hung Hom}
 \country{Hong Kong}}
\email{haobin.ke@connect.polyu.hk}

\author{Sen Zhang}
\affiliation{%
 \institution{The Hong Kong Polytechnic University}
 \city{Hung Hom}
 \country{Hong Kong}}
\email{senzhang@polyu.edu.hk}

\author{Qingqing Ye}
\affiliation{%
 \institution{The Hong Kong Polytechnic University}
 \city{Hung Hom}
 \country{Hong Kong}}
\email{qqing.ye@polyu.edu.hk}

\author{Xun Ran}
\affiliation{%
 \institution{The Hong Kong Polytechnic University}
 \city{Hung Hom}
 \country{Hong Kong}}
\email{qi-xun.ran@connect.polyu.hk}

\author{Haibo Hu}
\authornote{Corresponding author}
\affiliation{%
 \institution{The Hong Kong Polytechnic University}
\institution{Research Centre for Privacy and Security Technologies in Future Smart Systems, PolyU}
 \city{Hung Hom}
 \country{Hong Kong}}
\email{haibo.hu@polyu.edu.hk}


\begin{abstract}
Signed graphs with positive and negative edges can model complex relationships in social networks. Leveraging on balance theory that deduces edge signs from multi-hop node pairs, signed graph learning can generate node embeddings that preserve both structural and sign information.
However, training on sensitive signed graphs raises significant privacy concerns, as model parameters may leak private link information. Existing protection methods with differential privacy (DP) typically rely on edge or gradient perturbation for unsigned graph protection. Yet, they are not well-suited for signed graphs, mainly because edge perturbation tends to cascading errors in edge sign inference under balance theory, while gradient perturbation increases sensitivity due to node interdependence and gradient polarity change caused by sign flips, resulting in larger noise injection. In this paper, motivated by the robustness of adversarial learning to noisy interactions, we present ASGL, a privacy-preserving adversarial signed graph learning method that preserves high utility while achieving node-level DP. We first decompose signed graphs into positive and negative subgraphs based on edge signs, and then design a gradient-perturbed adversarial module to approximate the true signed connectivity distribution. In particular, the gradient perturbation helps mitigate cascading errors, while the subgraph separation facilitates sensitivity reduction. Further, we devise a constrained breadth-first search tree strategy that fuses with balance theory to identify the edge signs between generated node pairs. This strategy also enables gradient decoupling, thereby effectively lowering gradient sensitivity. Extensive experiments on real-world datasets show that ASGL achieves favorable privacy-utility trade-offs across multiple downstream tasks.
Our code and data are available in \textcolor{blue}{\url{https://github.com/KHBDL/ASGL-KDD26}}.
\end{abstract}





\maketitle
\vspace{-6pt}
\section{Introduction}\label{sec:01}
\begin{figure}[!t]
\setlength{\abovecaptionskip}{2pt}
\setlength{\belowcaptionskip}{-5pt}
  \centering
  \includegraphics[width=0.80\linewidth]{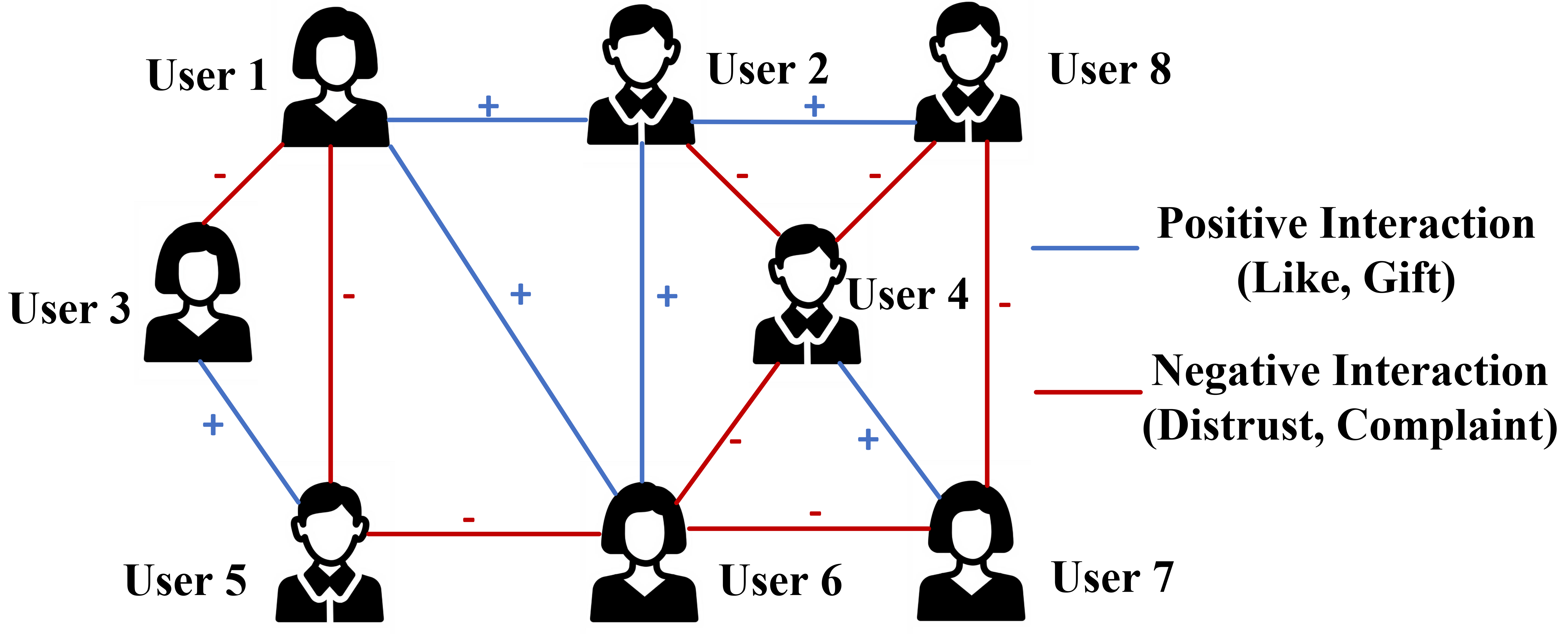}
  \caption{A signed social graph with blue edges for positive links and red edges for negative links.}
  \label{fig:f1}
\end{figure}

The signed graph is a common and widely adopted graph structure that can represent both positive and negative relationships using signed edges~\cite{19,20,21}. For example, in online social networks shown in Fig.~\ref{fig:f1}, while user interactions reflect positive relationships (e.g., like, trust, friendship), negative relationships (e.g., dislike, distrust, complaint) also exist. Signed graphs provide more expressive power than unsigned graphs to capture such complex user interactions.

Recently, some studies~\cite{22,23, 24} have explored signed graph learning methods, aiming to obtain low-dimensional vector representations of nodes that preserve key signed graph properties: neighbor proximity and structural balance. These embeddings are subsequently applied to downstream tasks such as edge sign prediction, node clustering, and node classification. 
Among existing signed graph learning methods, \textit{balance theory}~\cite{27} has proven effective in identifying the edge signs between the source node and multi-hop neighbor nodes. 
It is leveraged in graph neural network (GNN)-based models to guide message passing across signed edges, ensuring that information aggregation is aligned with the node proximity~\cite{36,38,39}. Moreover, to enhance the robustness and generalization capability of deep learning models, the adversarial graph embedding model~\cite{03,14} learns the underlying connectivity distribution of signed graphs by generating high-quality node embeddings that preserve signed node proximity. 

Despite their ability to effectively capture signed relationships between nodes, graph learning models remain vulnerable to link stealing attacks~\cite{25,42,43}, which aim to infer the existence of links between arbitrary node pairs in the training graph. For instance, in online social graphs, such attacks may reveal whether two users share a friendly or adversarial relationship, compromising user privacy and damaging personal or professional reputations.

Differential privacy (DP)~\cite{06} is a rigorous privacy framework that guarantees statistically indistinguishable outputs regardless of any individual data presence. Such guarantee is achieved through sufficient perturbation while maintaining provable privacy bounds and computational feasibility.
Existing privacy-preserving graph learning methods with DP can be categorized into two types based on the perturbation mechanism: one applies edge perturbation~\cite{53} to protect the link information by modifying the graph structure, and the other adopts gradient perturbation~\cite{54,52} to obscure the relationships between nodes during model training. However, these methods are not well-suited for signed graph learning due to the following two challenges:
\begin{itemize}[leftmargin=5mm]
\item \textbf{\textit{Cascading error}}: As illustrated in Fig.~\ref{fig:f2}, balance theory facilitates the inference of the edge sign between two unconnected nodes by computing the product of edge signs along a path. However, existing methods that use edge perturbation to protect link information may alter the sign of any edge along the path, thereby leading to incorrect inference of edge signs under balance theory. Such a local error can further propagate along the path, resulting in cascading errors in edge sign inference.
\item \textbf{\textit{High sensitivity}}:
While gradient perturbation methods without directly perturbing edges may mitigate cascading errors, they are still ill-suited for signed graph learning because the node interdependence in signed graphs leads to high gradient sensitivity.\footnote{The presence or absence of a node affects gradient updates of itself and its neighbors.} Furthermore, edge change may induce sign flips that reverse gradient polarity within the loss function (see Eq.~(\ref{eq: loss_t}) for details), resulting in higher sensitivity compared to unsigned graphs. This increased sensitivity requires larger noise for privacy protection, thereby reducing the data utility.
\end{itemize}

To address these challenges, we turn to an adversarial learning-based approach for private signed graph learning. The core motivation is that this adversarial method generates node embeddings by approximating the true connectivity distribution, making it naturally robust to noisy interactions during optimization. As a result, we propose ASGL, a differentially private adversarial signed graph learning method that achieves high utility while maintaining node-level differential privacy. Within ASGL, the signed graph is first decomposed into positive and negative subgraphs based on edge signs. These subgraphs are then processed through an adversarial learning module within shared model parameters, enabling both positive and negative node pairs to be mapped into a unified embedding space while effectively preserving signed proximity.
Based on this, we develop the adversarial learning module with differentially private stochastic gradient descent (DPSGD), which generates private node embeddings that closely approximate the true signed connectivity distribution. In particular, \emph{the gradient perturbation helps mitigate cascading errors, while the subgraph separation avoids gradient polarity reversals induced by edge sign flips within the loss function, thereby reducing the sensitivity to changes in edge signs.}
Considering that node interdependence further increases gradient sensitivity, we design a constrained breadth-first search (BFS) tree strategy within adversarial learning. \emph{This strategy integrates balance theory to identify the edge signs between generated node pairs, while also constraining the receptive fields of nodes to enable gradient decoupling, thereby effectively lowering gradient sensitivity and reducing noise injection.} Our main contributions are listed as follows:
\begin{itemize}[leftmargin=5mm]
\item
We present a privacy-preserving adversarial learning method for signed graphs, called ASGL. To our best knowledge, it is the \textbf{\textit{first}} work that can ensure the node-level differential privacy of signed graph learning while preserving high data utility.
\item To mitigate cascading errors, we develop the adversarial learning module with DPSGD, which generates private node embeddings that closely approximate the true signed connectivity distribution. This approach avoids direct perturbation of the edge structure, which helps mitigate cascading errors and prevents gradient polarity reversals in the loss function.
\item 
To further reduce the sensitivity caused by complex node relationships, we design a constrained breadth-first search tree strategy that integrates balance theory to identify edge signs between generated node pairs. This strategy also constrains the receptive fields of nodes, enabling gradient decoupling and effectively lowering gradient sensitivity.
\item
Extensive experiments demonstrate that our method achieves favorable privacy-accuracy trade-offs and significantly outperforms state-of-the-art methods in edge sign prediction and node clustering tasks.
Additionally, we conduct link stealing attacks, demonstrating that ASGL exhibits stronger resistance to such attacks across all datasets.
\end{itemize}

The remainder of our work is organized as follows. Section~\ref{sec:preliminary} describes the preliminaries of our solution. The problem statement is introduced in Section~\ref{sec:Prob_form}. 
Our proposed solution and its privacy analysis are presented in Section~\ref{sec:ASGL_MainMethod}.
The experimental results are reported in Section~\ref{sec:experiments}. We discuss related works in Section~\ref{sec:related_work}, followed by conclusion in Section~\ref{sec:conclusion}.


\section{Preliminaries}\label{sec:preliminary}
In this section, we provide an overview of signed graphs, differential privacy, and DPSGD. Additionally, the vanilla adversarial graph learning is introduced in \textbf{App.~\ref{app:AL}}, and the frequently used notations are summarized in~\autoref{tab:01} (See  \textbf{App.~\ref{app:notations}}). 


\subsection{Signed Graph with Balance Theory}
\label{sec: signed graph}
A signed graph is denoted as $\mathcal{G}=(V,E^+,E^-)$, where $V$ is the set of nodes, and $E^+/E^-$ represent positive and negative edge sets, respectively.
An edge $e_{ij}=(v_i,v_j) \in E^+/E^-$ represents the positive/negative link between node pair $(v_i,v_j) \in V$, respectively. 
Notably, $E^+ \cap E^-= \emptyset$ ensures that any node pair cannot maintain both positive and negative relationships simultaneously. 
The objective of signed graph embedding is to learn a mapping function $f:V\rightarrow \mathbb{R}^k$ that projects each node $v \in V$ into a low $k$-dimensional vector while preserving both the structural properties of the original signed graph.
In other words,  node pairs connected by positive edges should be embedded closely, while those connected by negative edges should be placed farther apart in the embedding space.
\vspace{-5pt}
\begin{figure}[!h]
\setlength{\abovecaptionskip}{2pt}
\setlength{\belowcaptionskip}{-5pt}
  \centering
  \includegraphics[width=0.9\linewidth]{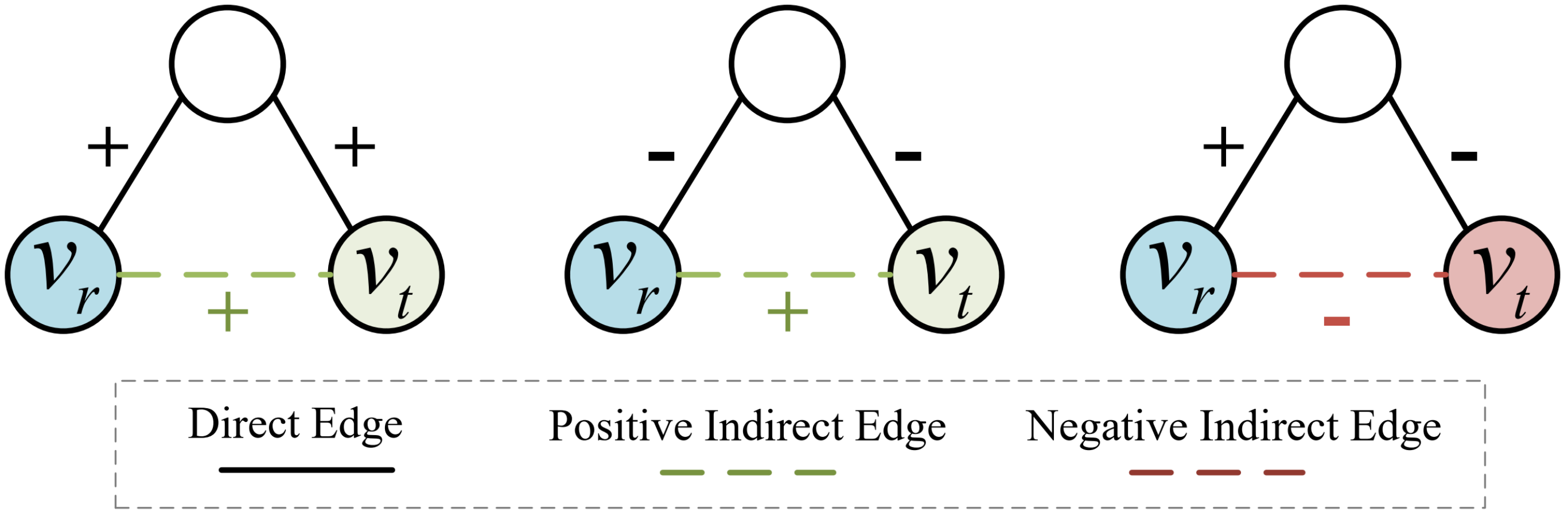}
  \caption{The signs of multi-hop connection based on balanced theory.}
  \label{fig:f2}
\end{figure}
\vspace{-5pt}

Balance theory~\cite{27} is a well-established standard to describe the signed relationships of unconnected node pairs. It is commonly summarized by four intuitive rules: “A friend of my friend is my friend,” “A friend of my enemy is my enemy,” “An enemy of my friend is my enemy,” and “An enemy of my enemy is my friend.” 
Based on these rules, the balance theory can deduce signs of the multi-hop connection. As shown in Fig.~\ref{fig:f2}, given a path $P_{rt}: v_r \rightarrow v_t$ from rooted node $v_r$ to target node $v_t$, the sign of the indirect relationships between $v_r$ and $v_t$ can be inferred by iteratively applying balance theory. Specifically, the sign of the multi-hop connection corresponds to the product of the signs of the edges along the path.

\subsection{Differential Privacy} 
Differential Privacy (DP)~\cite{04} provides a rigorous mathematical framework for quantifying the privacy guarantees of algorithms operating on sensitive data. Informally, it bounds how much the output distribution of a mechanism can change in response to small changes in its input.
When applying DP to signed graph data, the definition of adjacent databases typically considers two signed graphs, $\mathcal{G}$ and $\mathcal{G^\prime}$, which are regarded as adjacent graphs if they differ by at most one edge or one node with its associated edges.

\begin{definition}[Edge (Node)-level DP~\cite{05}]
Given $\epsilon> 0$ and $\delta > 0$, a graph analysis mechanism $\mathcal{M}$ satisfies edge- or node-level $(\epsilon,\delta)$-DP, if for any two adjacent graph datasets $\mathcal{G}$ and $\mathcal{G^\prime}$ that only differ by an edge or a node with its associated edges, and for any possible algorithm output $S \subseteq Range(\mathcal{M})$, it holds that
\begin{align}
\text{Pr}[\mathcal{M}(\mathcal{G}) \in S] \leq e^{\epsilon} \text{Pr}[\mathcal{M}(\mathcal{G^\prime}) \in S] + \delta.
\end{align}
Here, $\epsilon$ is the privacy budget (i.e., privacy cost), where smaller values indicate stronger privacy protection but greater utility reduction. The parameter $\delta$ denotes the probability that the privacy guarantee may not hold, and is typically set to be negligible. In other words, $\delta$ allows for a negligible probability of privacy leakage, while ensuring the privacy guarantee holds with high probability.
\label{def:01}
\end{definition}

\begin{remark}
Note that satisfying node-level DP is much more challenging than satisfying edge-level DP, as removing a single node may, in the worst case, remove $|V|-1$ edges, where $|V|$ denotes the total number of nodes. Consequently, node-level DP requires injecting substantially more noise.
\end{remark}

Two fundamental properties of DP are useful for the privacy analysis of complex algorithms:
(1) \textbf{Post-Processing Property}~\cite{06}: If a mechanism $\mathcal{M}(\mathcal{G})$ satisfies \((\epsilon, \delta)\)-DP, then for any function \( f \) that indirectly queries the private dataset \( \mathcal{G} \), the composition \( f(\mathcal{M}(\mathcal{G})) \) also satisfies \((\epsilon, \delta)\)-DP;
(2) \textbf{Composition Property}~\cite{06}: If $\mathcal{M}(\mathcal{G})$ and $f(\mathcal{G})$ satisfy \((\epsilon_1, \delta_1)\)-DP and \((\epsilon_2, \delta_2)\)-DP, respectively, then the combined mechanism $\mathcal{F}(\mathcal{G})=(\mathcal{M}(\mathcal{G}),f(\mathcal{G}))$ which outputs both results, satisfies \((\epsilon_1+\epsilon_2, \delta_1+\delta_2)\)-DP.

\subsection{DPSGD} 
\label{sub:DPSGD}
A common approach to differentially private training combines noisy stochastic gradient descent with the Moments Accountant (MA)~\cite{02}. This approach, known as DPSGD, has been widely adopted for releasing private low-dimensional representations, as MA effectively mitigates excessive privacy loss during iterative optimization. Formally, for each sample $x_i$ in a batch of size $B$, we compute its gradient $\nabla \mathcal{L}_i(\theta)$, denoted as $\nabla(x_i) $ for simplicity. 
\textbf{Gradient sensitivity} refers to the maximum change in the output of the gradient function resulting from a change in a single sample.
To control the sensitivity of ${\nabla(x_i)}$, the $\ell_2$ norm of each gradient is clipped by a threshold $C$. These clipped gradients are then aggregated and perturbed with Gaussian noise $\mathcal{N}(0,\sigma^2 C^2\mathbf{I})$ to satisfy the DP guarantee. Finally, the average noisy gradient ${\tilde{\nabla}_B}$ is used to update the model parameters $\theta$.
This process is given by:
\begin{align}
{\tilde{\nabla}_B} \leftarrow \frac{1}{B} \Big(\sum_{i=1}^B\text{Clip}_C(\nabla(x_i))+\mathcal{N}\left(0, \sigma^2 C^2 \mathbf{I}\right)\Big).
\label{eq:dpsgd}
\end{align}
Here, $\text{Clip}_C(\nabla(x_i)) = \nabla(x_i)/\max(1,\frac{||\nabla(x_i)||_2}{C})$. 
\section{Problem Definition and Existing Solutions}\label{sec:Prob_form}
\subsection{Problem Definition}
Instead of publishing a sanitized version of original node embeddings, we aim to release a privacy-preserving ASGL model trained on raw signed graph data with node-level DP guarantees, enabling data analysts to generate task-specific node embeddings.

\textbf{Threat Model}. 
We consider a black-box attack~\cite{42}, where the attacker can query the trained model and observe its outputs with no access to its internal architecture or parameters. The attacker attempts to infer the presence of specific nodes or edges in the training graph solely from model outputs. This setting reflects a more practical attack surface compared to the white-box scenario~\cite{11}.

\textbf{Privacy Model.} 
Signed graph data encodes both positive and negative relationships between nodes, which differs from tabular or image data. Therefore, it is necessary to adapt the standard definition of node-level DP (See Definition~\ref{def:01}) to ensure black-box adversaries cannot determine whether a specific node and its associated signed edges are present in the training data. To this end, we define the differentially private adversarial signed graph learning model as follows.

\begin{definition}[Adversarial signed graph learning model under node-level DP]
\label{def:04}
The vanilla process of graph adversarial learning is illustrated in  \textbf{App.~\ref{app:AL}}, let $\theta_D$ denote the discriminator parameters, and its $r$-th row element corresponds to the $k$-dimensional vector $\mathbf{d}_{v_r}$ of node $v_r$, that is $\mathbf{d}_{v_r} \in \theta_D$. The discriminator module $L_D$ satisfies node-level ($\epsilon,\delta$)-DP if two adjacent signed graphs $\mathcal{G}$ and $\mathcal{G}^\prime$ only differ in one node with its associated signed edges, and for all possible $\theta_s \subseteq Range(L_D)$, we have
\begin{align}
\text{Pr}[L_D(\mathcal{G}) \in \theta_s] \leq e^{\epsilon} \text{Pr}[L_D(\mathcal{G^\prime}) \in \theta_s] + \delta,
\end{align}
where $\theta_s$ denotes the set comprising all possible values of $\theta_D$.
\end{definition}

In particular, the generator $G$ is trained based on the feedback from the differentially private discriminator $D$. According to the post-processing property of DP~\cite{08,12}, the generator module $L_G$ also satisfies node-level $(\epsilon, \delta)$-DP. Leveraging the robustness to post-processing property, the privacy guarantee is preserved in the generated signed node embeddings and their downstream usage.

\begin{figure*}[!t]
\setlength{\abovecaptionskip}{2pt}
\setlength{\belowcaptionskip}{-5pt}
  \centering
  \includegraphics[width=0.87\linewidth]{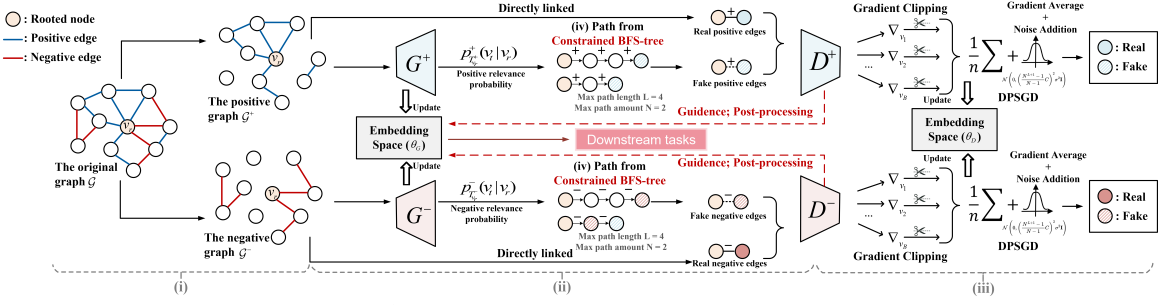}
  \caption{Overview of the ASGL framework: (i) The process decomposes a signed graph into positive and negative subgraphs, (ii) then maps node pairs into a unified embedding space while preserving signed proximity. To ensure privacy, (iii) adversarial learning module with DPSGD generates private node embeddings that approximate true connectivity without cascading errors. (iv) A constrained BFS-tree strategy manages node receptive field, reduces gradient noise, and improves model utility.}

  \label{fig:f3}
\end{figure*}

\subsection{Existing Solutions}
\label{sec: naive solution}
To our best knowledge, existing differentially private graph learning methods follow two main tracks: \textbf{gradient perturbation} and \textbf{edge perturbation}.
In the first category, Yang et al.~\cite{54} introduce a privacy-preserving generative model that incorporates generative adversarial networks (GAN) or variational autoencoders (VAE) with DPSGD to protect edge privacy, while Xiang et al.~\cite{52} design a node sampling mechanism that adds Laplace noise to per-subgraph gradients, achieving node-level DP.
For the edge perturbation-based methods, Lin et al.~\cite{53} use randomized response to perturb the adjacency matrix for edge-level privacy, and EDGERAND~\cite{42} perturbs the graph structure while preserving sparsity by clipping the adjacency matrix according to a privacy-calibrated graph density.

\textbf{Limitation}. 
The aforementioned solutions are not directly applicable to signed graphs. This is primarily because edge perturbation can lead to cascading errors when inferring edge signs under balance theory. 
Moreover, gradient perturbation often suffers from high sensitivity caused by complex node dependencies and gradient polarity reversal from edge sign flips, leading to excessive noise and degraded model utility.

\section{Our Proposal: ASGL}\label{sec:ASGL_MainMethod}
To tackle the above limitations, we present ASGL, a DP-based adversarial signed graph learning model that integrates a constrained BFS-tree strategy to achieve favorable utility-privacy tradeoffs.

\subsection{Overview}
The ASGL framework, illustrated in Fig.~\ref{fig:f3}, comprises three steps:

\begin{itemize}[leftmargin=5mm]
\item 
\textbf{Private Adversarial Signed Graph Learning.} The signed graph $\mathcal{G}$ is first split into positive and negative subgraphs, $\mathcal{G}^+$ and $\mathcal{G}^-$, based on edge signs.
Subsequently, two discriminators, $D^+$ and $D^-$, sharing parameters $\theta_D$, are trained to distinguish real from fake positive and negative edges.
Guided by $D^+$ and $D^-$, two generators $G^+$ and $G^-$ with shared parameters $\theta_G$ generate node embeddings that approximate the true connectivity distribution. 
To ensure node-level DP, we apply gradient perturbation during discriminator training instead of directly perturbing edges. This strategy mitigates cascading errors and prevents gradient polarity reversals caused by edge sign flips, thereby reducing gradient sensitivity. By the post-processing property, the generators also preserve node-level DP. 
\item \textbf{Optimization via Constrained BFS-tree.}
To further reduce gradient sensitivity and the required noise scale, ASGL employs a constrained BFS-tree strategy. By empirically limiting the number and length of paths, each node’s receptive field is restricted, which reduces node dependency and enables gradient decoupling. This significantly lowers gradient sensitivity and enhances model utility under differential privacy constraints. 
\item \textbf{Privacy Accounting and Complexity Analysis.}
The complete training process for ASGL is outlined in Algorithm~\ref{algo:ASGL} (see \textbf{App.~\ref{sec:training}}). 
Based on this, we present a comprehensive privacy accounting and computational complexity analysis for ASGL.  
\end{itemize}

\subsection{Private Adversarial Signed Graph Learning}
\label{sub:Optimization}
Motivated by~\cite{03,14}, a signed graph $\mathcal{G}$ is first divided into a positive subgraph $\mathcal{G}^+$ and a negative subgraph $\mathcal{G}^-$ according to edge signs. Let $\mathcal{N}(v_r)$ be the set of neighbor nodes directly connected to node $v_r$. We denote the true positive and negative connectivity distributions of $v_r$ over its neighborhood $\mathcal{N}(v_r)$ as the conditional probabilities $p_{\text{true}}^{+}(\cdot | v_r)$ and $p_{\text{true}}^{-}(\cdot | v_r)$, which capture the preference of $v_r$ to connect with other nodes in $V$. The adversarial learning for the signed graph $\mathcal{G}$ is conducted by two adversarial learning modules:

\textbf{Generators $G^+$ and $G^-$}: Through optimizing the shared parameters $\theta_G$, generators $G^+$ and $G^-$ aim to approximate the underlying true connectivity distribution and generate the most likely but unconnected nodes $v_t \notin \mathcal{N}(v_r)$ that are relevant to a given node $v_r$. To this end, we estimate the relevance probabilities of these \textbf{\textit{fake}}\footnote{The term “Fake” indicates that although a node $v$ selected by the generator is relevant to $v_r$, there is no actual edge between them.} node pairs. Specifically, for the implementation of $G^+$,  given the fake positive node pairs $(v_r,v_t)^+$, we use the graph softmax function~\cite{03} to calculate the fake positive connectivity probability:
\begin{equation}
\small
\label{eq: FPCP}
p^+_{\text{fake}}(v_t| v_r)= G^+\left(v_t| v_r;\theta_G\right) = \sigma(\mathbf{g}_{v_t}^\top  \mathbf{g}_{v_r}) =\frac{1}{1+\exp({-\mathbf{g}_{v_t}^\top  \mathbf{g}_{v_r})}},
\end{equation}
where $\mathbf{g}_{v_t}, \mathbf{g}_{v_r} \in \mathbb{R}^k$ are the $k$-dimensional vectors of nodes $v_t$ and $v_r$, respectively, and $\theta_G$ is the union of all $\mathbf{g}_v$'s. The output $G^+(v_t|v_r;\theta_G)$ increases with the decrease of the distance between $v_r$ and $v_t$ in the embedding space of the generator $G^+$. 
Similarly, for the generator $G^-$, given the fake negative node pairs $(v_r,v_t)^-$, we estimate their fake negative connectivity probability:
\begin{equation}
\small
\label{eq: FNCP}
p^-_{\text{fake}}(v_t|v_r) = G^-(v_t|v_r;\theta_G) = 1 - \sigma(\mathbf{g}_{v_t}^\top\!\mathbf{g}_{v_r}) = \frac{\exp{(-\mathbf{g}_{v_t}^\top\!\mathbf{g}_{v_r}})}{1 + \exp{(-\mathbf{g}_{v_t}^\top\!\mathbf{g}_{v_r}})}.
\end{equation}
Here, Eq.~\eqref{eq: FNCP} ensures that node pairs with higher negative connectivity probabilities are mapped farther apart in the embedding space of $G^-$. Since generators $G^+$ and $G^-$ share the parameters $\theta_G$, they jointly learn the proximity and separation of positive and negative node pairs in a unified embedding space, respectively.

Notably, the aforementioned fake node pairs $(v_r,v_t)^+$ and $(v_r,v_t)^-$ are sampled by a breadth-first search (BFS)-tree strategy~\cite{27}. Compared to depth-first search (DFS)~\cite{56}, BFS ensures more uniform exploration of neighboring nodes and can be integrated with random walk techniques~\cite{29} to optimize computational efficiency.
Specifically, we perform BFS on the positive subgraph $\mathcal{G}^+$ to construct a BFS-tree $T^+_{v_r}$ rooted from node $v_r$. Then, we calculate the positive relevance probability of node $v_r$ with its neighbors $v_k \in \mathcal{N}({v_r})$:
\begin{equation}
\small
\label{eq:11}
p^+_{T^+_{v_r}}(v_k | v_r)=\frac{\exp \left(\mathbf{g}_{v_k}^{\top} \mathbf{g}_{v_r}\right)}{\sum_{v_k  \in \mathcal{N}({v_r})} \exp \left(\mathbf{g}_{v_k}^{\top} \mathbf{g}_{v_r}\right)},
\end{equation}
which is actually a softmax function over $\mathcal{N}({v_r})$.
To further sample node pairs unconnected in $T^+_{v_r}$ as fake positive edges, we perform a random walk at $T^+_{v_r}$:     
Starting from the root node $v_r$, a path $P_{rt}: v_r \rightarrow v_t$ is built by iteratively selecting the next node based on the transition probabilities defined in Eq.~\eqref{eq:11}. The resulting unconnected node pair $(v_r, v_t)^+$ is treated as a fake positive edge, and  \textbf{App.~\ref{sec:BFS}} provides an example of this process.
Given the node pair $(v_{r}, v_{t})^+$,  the generator $G^+$ estimates $p^+_{\text{fake}}(v_t |v_r)$ according to Eq.~\eqref{eq: FPCP}.

Similarly, we also establish a BFS-tree $T^-_{v_r}$ rooted at node $v_r$ in the negative subgraph $\mathcal{G}^-$. To obtain the negative node pair $(v_r,v_t)^-$, we perform a random walk on $T^-_{v_r}$ according to the following transition probability (i.e., negative relevance probability):
\begin{equation}
\small
\label{eq:13}
p^-_{T^-_{v_r}}(v_k | v_r)=\frac{1-\exp \left(\mathbf{g}_{v_k}^{\top} \mathbf{g}_{v_r}\right)}{\sum_{v_k  \in \mathcal{N}({v_r})}\left(1-\exp \left(\mathbf{g}_{v_k}^{\top} \mathbf{g}_{v_r}\right)\right)}.
\end{equation}

In particular, the edge sign of the negative node pair $(v_r, v_t)^-$ depends on the length of the path $P_{rt}: v_r \rightarrow v_t$. According to the balance theory introduced in Section~\ref{sec: signed graph}, the edge signs of multi-hop node pairs correspond to the product of the edge signs along the path. Accordingly, the rules for generating fake negative edges within $P_{rt}$ are defined as follows:
(1) If the path length of $P_{rt}$ is odd, a node pair $(v_r,v_t)^-$ for the rooted node $v_r$ and the last node $v_t$ is selected as a fake negative pair;
(2) If the path length of $P_{rt}$ is even, a node pair $(v_r,v_t)^-$ for the rooted node $v_r$ and the second last node $v_t$ is selected as a fake negative pair.
The resulting node pair $(v_r, v_t)^-$ is then used to compute $p^-_{\text{fake}}(v_t | v_r)$ according to Eq.~\eqref{eq: FNCP}.

\textbf{Discriminators $D^+$ and $D^-$}: This module tries to distinguish between real node pairs and fake node pairs synthesized by the generators $G^+$ and $G^-$. Accordingly, the discriminators $D^+$ and $D^-$ estimate the likelihood that positive and negative edges exists between $v_r$ and $v\in V$, respectively, denoted as:
\begin{equation}
\small
\label{eq:D+}
D^+(v_r,v|\theta_D) = \sigma(\mathbf{d}_{v}^\top  \mathbf{d}_{v_r}) = \frac{1}{1+\exp({-\mathbf{d}_{v}^\top  \mathbf{d}_{v_r})}},\\
\end{equation}
\vspace{-0.7em}
\begin{equation}
\small
\label{eq:D-}
D^-(v,v_r|\theta_D) = 1-\sigma(\mathbf{d}_{v}^\top  \mathbf{d}_{v_r}) = \frac{\exp({-\mathbf{d}_{v}^\top  \mathbf{d}_{v_r})}}{1+\exp({-\mathbf{d}_{v}^\top  \mathbf{d}_{v_r})}},
\end{equation}
where $\mathbf{d}_v, \mathbf{d}_{v_r} \in \mathbb{R}^k$ are vectors corresponding to the $v$-th and $v_r$-th rows of shared parameters $\theta_D$, respectively. $\sigma(\cdot)$ represents the sigmoid function of the inner product of these two vectors. 

In summary, given real positive and real negative edges sampled from $p_{\text{true}}^{+}(\cdot | v_r)$ and $p_{\text{true}}^{-}(\cdot | v_r)$, along with fake positive and fake negative edges generated from generators $G^+/G^-$, the adversarial learning pairs $(D^+, G^+)$ and $(D^-, G^-)$, operating on the positive subgraph $\mathcal{G}^+$ and the negative subgraph $\mathcal{G}^-$, respectively, engage in a four-player mini-max game with the joint loss function: 
\begin{equation}
\small
\label{eq: loss_t}
\begin{aligned}
\min _{\theta_G}&  \max _{\theta_D} L \left(G^{+}, G^{-}, D^{+}, D^{-}\right) \\
= & \sum_{v_r \in V^{+}}\left(\left(\mathbb{E}_{v \sim p_{\text {true }}^{+}\left(\cdot \mid v_r\right)}\right)\left[\log D^{+}\left(v, v_r \mid \theta_D\right)\right]\right. \\
& \left.\quad+\left(\mathbb{E}_{v \sim G^{+}\left(\cdot \mid v_r ; \theta_G\right)}\right)\left[\log \left(1-D^{+}\left(v, v_r \mid \theta_D\right)\right)\right]\right) \\
+ & \sum_{v_r \in V^{-}}\left(\left(\mathbb{E}_{v \sim p_{\text {true }}^{-}\left(\cdot \mid v_r\right)}\right)\left[\log D^{-}\left(v, v_r \mid \theta_D\right)\right]\right. \\
& \left.\quad+\left(\mathbb{E}_{v \sim G^{-}\left(\cdot \mid v_r ; \theta_G\right)}\right)\left[\log \left(1-D^{-}\left(v, v_r \mid \theta_D\right)\right)\right]\right) .
\end{aligned}
\end{equation}
Based on Eq.~\eqref{eq: loss_t}, the parameters $\theta_D$ and $\theta_G$ are updated alternately by maximizing and minimizing the joint loss function. Competition between $G$ and $D$ results in mutual improvement until the fake node pairs generated by $G$ are indistinguishable from the real ones, thus approximating the true connectivity distribution. Lastly, the learned node embeddings $\mathbf{g}_v \in \theta_G$ are used in downstream tasks.

\emph{\textbf{How to Achieve DP?}} Given real and fake positive/negative edges of the node $v_i$, the corresponding node embedding $\mathbf{d}_{v_i} \in \theta_D$ is updated by ascending gradients of the joint loss function in Eq.~\eqref{eq: loss_t}:
\begin{equation}
\label{eq: LD_all}
\small
\frac{\partial L_D}{\partial \mathbf{d}_{v_i}} =\left\{\begin{array}{l}
\partial \log{D^+(v_i,v_j|\theta_D)}/{\partial \mathbf{d}_{v_i}}=[1-\sigma(\mathbf{d}_{v_j}^\top  \mathbf{d}_{v_i})]\mathbf{d}_{v_j},\\
\text {if }\left(v_i, v_j\right) \text{ is a real positive edge from $\mathcal{G}^+$}; \\ 
\partial \log{(1-D^+(v_i,v_j|\theta_D))}/{\partial \mathbf{d}_{v_i}}=-\sigma(\mathbf{d}_{v_j}^\top  \mathbf{d}_{v_i}) \mathbf{d}_{v_j}, \\
\text {if }\left(v_i, v_j\right)\text{ is a fake positive edge from ${G}^+$}; \\ 
\partial \log{D^-(v_i,v_j|\theta_D)}/{\partial \mathbf{d}_{v_i}}=-\sigma(\mathbf{d}_{v_j}^\top  \mathbf{d}_{v_i})\mathbf{d}_{v_j},\\
\text {if }\left(v_i, v_j\right) \text{ is a real negative edge from $\mathcal{G}^-$}; \\ 
\partial \log{(1-D^-(v_i,v_j|\theta_D))}/{\partial \mathbf{d}_{v_i}}=[1-\sigma(\mathbf{d}_{v_j}^\top  \mathbf{d}_{v_i})]\mathbf{d}_{v_j}, \\
\text {if }\left(v_i, v_j\right)\text{ is a fake negative edge from ${G}^-$}. 
\end{array}\right.
\end{equation}

According to Definition~\ref{def:04}, to achieve node-level differential privacy in adversarial signed graph learning, it is necessary to add the Gaussian noise to the sum of clipped gradients over a batch of nodes. The resulting noisy gradient $\tilde{\nabla}{L_D}$ is formulated as:
\begin{equation}
\small
\label{eq: L_D}
{\tilde{\nabla}{L_D}} = \frac{1}{B} \Big(\sum_{v_i \in V_B}\text{Clip}_C(\frac{\partial L_D}{\partial \mathbf{d}_{v_i}})+\mathcal{N}\left(0, B^2 C^2\sigma^2 \mathbf{I}\right)\Big),
\end{equation}
where $V_B$ denotes the batch set of nodes, with batch size $B = |V_B|$. $C$ is the clipping threshold to control gradient sensitivity. The fact that the gradient sensitivity reaches $BC$ is explained in Section~\ref{sub:Constrained BFS-trees}.
\begin{remark}
To achieve node-level DP, we perturb discriminator gradients instead of signed edges, avoiding cascading errors and gradient polarity reversals from edge sign flips (see Eq.~(\ref{eq: loss_t})), which reduces gradient sensitivity. Furthermore, generators also preserve DP under discriminator guidance via the post-processing property of DP.
\end{remark}

\subsection{Optimization via Constrained BFS-Tree}
\label{sub:Constrained BFS-trees}
\begin{figure}[!t]
\setlength{\abovecaptionskip}{2pt}
\setlength{\belowcaptionskip}{-5pt}
  \centering
  \includegraphics[width=0.7\linewidth]{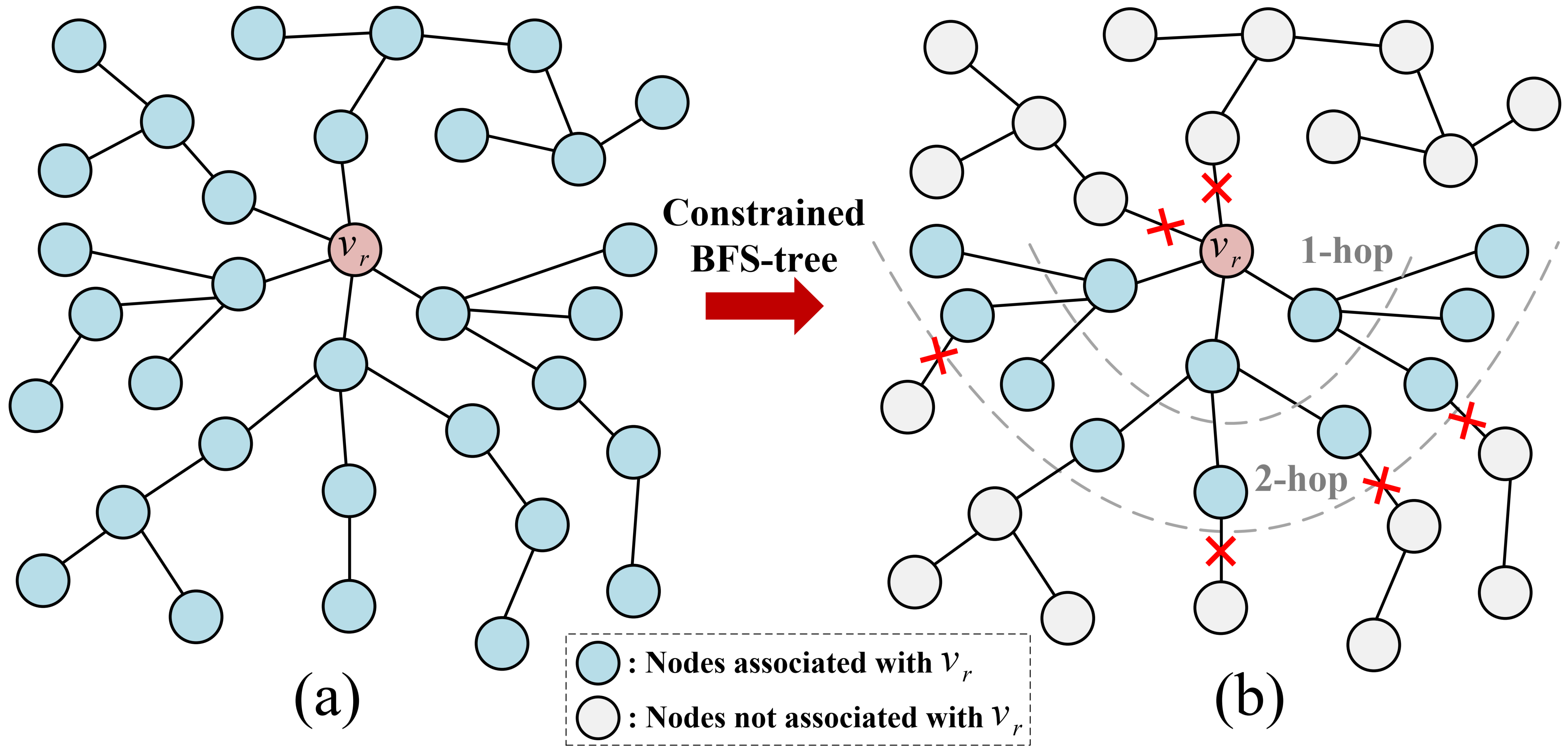}
  \caption{The receptive field of node $v_r$ within a batch is illustrated in two cases: (a) An unconstrained BFS tree, and the receptive field size of $v_r$ is $B=|V_B|=34$; (b) A constrained BFS tree with path length $L=2$, path amount $N=3$ of each node, and the receptive field size of $v_r$ is $\sum_{l=0}^L N^l=13$.}
  \label{fig:receptive field}
\end{figure}

According to Eq.~(\ref{eq: LD_all}), in graph adversarial learning, the interdependence among samples implies that modifying a single node $v_i$ may affect the gradients of multiple other nodes $v_j$ within the same batch. This interdependence also exists among the fake node pairs generated along the BFS-tree paths. Consequently, in the worst-case illustrated in Fig.~\ref{fig:receptive field}(a), all node samples within a batch may become interrelated due to the BFS-tree, resulting in the gradient sensitivity of discriminators $D$ as high as $BC$. Such high sensitivity necessitates injecting substantial noise to satisfy node-level DP, hindering effective optimization and reducing model utility.

To address the aforementioned challenge, we introduce the \textbf{\textit{constrained BFS-tree strategy}}: As illustrated in Algorithm~\ref{algo:SAMPLE-SUBGRAPHS} (see \textbf{App.~\ref{sec:BFS-TREE}}), when performing a random walk on the BFS-tree $T^+_{v_r}$ or $T^-_{v_r}$ rooted at $v_r \in V_{tr}$ to generate multiple unique paths, we also limit both the number of sampled paths and their lengths by $N$ and $L$. Following this, the training set of subgraphs $S_{tr}$ composed of constrained paths is obtained. The rationale behind these settings is discussed below.

\begin{theorem}
\label{theorem:L_D_NEW}
By constraining both the number and length of paths generated via random walks on the BFS-trees to $N$ and $L$, respectively, the gradient sensitivity $\Delta_{{g}}$ of the discriminator can be reduced from $BC$ to $\frac{N^{L+1}-1}{N-1}C$. \footnote{Empirical results in Section~\ref{sec:experiments} demonstrate that our ASGL achieves satisfactory performance even with a relatively small receptive field. Specifically, when setting $N = 3$ and $L = 4$, that is, $\frac{N^{L+1}-1}{N-1}=121<B=256$, the ASGL method still performs good model utility.} Thus, the noisy gradient $\tilde{\nabla}{L_D}$ of discriminator within a mini-batch $\mathcal{B}_{t}$ is denoted as:
\begin{align}
\label{eq:L_D_NEW}
{\tilde{\nabla}{L_D}} = \frac{1}{|\mathcal{B}_{t}|} \Big(\sum_{v \in \mathcal{B}_{t}}\text{Clip}_C(\frac{\partial L_D}{\partial \mathbf{d}_{v}})+\mathcal{N}\left(0, \Delta_{{g}}^2\sigma^2 \mathbf{I}\right)\Big),
\end{align}
where the gradient sensitivity $\Delta_{{g}}=\frac{N^{L+1}-1}{N-1}C$.
\end{theorem}

\noindent \textit{Proof of Theorem~\ref{theorem:L_D_NEW}}.
Let the sum of clipped gradients of batch subgraphs be $g_t(\mathcal{G}) =\sum_{v \in \mathcal{B}_{t}}\text{Clip}_C(\frac{\partial L_D}{\partial \mathbf{d}_v})$, where $\mathcal{B}_{t}$ represents any choice of batch subgraphs from $S_{tr}$. Consider a node-level adjacent graph $\mathcal{G}^\prime$ formed by removing a node $v^*$ with its associated edges from $\mathcal{G}$, we obtain their training sets of subgraphs $S_{tr}$ and $S_{tr}^{\prime}$ via the SAMPLE-SUBGRAPHS method in Algorithm~\ref{algo:SAMPLE-SUBGRAPHS}, denoted as:
\begin{equation}
\begin{aligned}
S_{tr}&=\text{SAMPLE-SUBGRAPHS}(\mathcal{G},V_{tr}, N, L),\\
S_{tr}^\prime&=\text{SAMPLE-SUBGRAPHS}(\mathcal{G}^\prime,V_{tr}, N, L).
\end{aligned}
\end{equation}

The only subgraphs that differ between $S_{tr}$ and $S_{tr}^\prime$ are those that involve the node $v^*$. Let $S(v^*)$ denote the set of such subgraphs, i.e., $S(v^*) = S_{tr} \setminus S_{tr}^\prime$. According to Lemma~\ref{lemma: Receptive field} in  \textbf{App.~\ref{app:lemma}}, the number of such subgraphs $S(v^*)$ is at most $R_{N,L}$. 
Thus, in any mini-batch training, the only gradient terms $\frac{\partial L_D}{\partial \mathbf{d}_v}$ affected by the removal of node $v^*$ are those associated with the subgraphs in $(S(v^*)\cap \mathcal{B}_{t})$:
\begin{align}
g_t(\mathcal{G}) - g_t(\mathcal{G}^\prime) 
&=\sum_{v \in \mathcal{B}_{t}}\text{Clip}_C(\frac{\partial L_D}{\partial \mathbf{d}_v}) - \sum_{v^\prime \in \mathcal{B}_{t}^\prime}\text{Clip}_C(\frac{\partial L_D}{\partial \mathbf{d}_{v^\prime}}) \\
&=\sum_{v,v^\prime \in (S(v^*)\cap \mathcal{B}_{t})}[\text{Clip}_C(\frac{\partial L_D}{\partial \mathbf{d}_v}) - \text{Clip}_C(\frac{\partial L_D}{\partial \mathbf{d}_{v^\prime}})], \notag
\end{align}
where $\mathcal{B}_{t}^\prime=\mathcal{B}_{t}\setminus (S(v^*)\cap \mathcal{B}_{t})$. Since each gradient term is clipped to have an $\ell_2$-norm of at most $C$, it holds that:
\begin{equation}
||\text{Clip}_C(\frac{\partial L_D}{\partial \mathbf{d}_v}) - \text{Clip}_C(\frac{\partial L_D}{\partial \mathbf{d}_{v^\prime}})||_F 
\leq C.
\end{equation}
In the worst case, all subgraphs in $S(v^*)$ appear in $\mathcal{B}_{t}$, so we bound the $\ell_2$-norm of the following quantity based on Lemma~\ref{lemma: S_v} in  \textbf{App.~\ref{app:lemma}}:
\begin{equation}
||g_t(\mathcal{G}) - g_t(\mathcal{G}^\prime) ||_F 
\leq C\cdot R_{N,L} 
= C\cdot\frac{N^{L+1}-1}{N-1}.
\end{equation}

The same reasoning applies when $\mathcal{G}^\prime$ is obtained by adding a new node $v^*$ to $\mathcal{G}$. Since $\mathcal{G}$ and $\mathcal{G}^\prime$ are arbitrary node-level adjacent graphs, the proof is complete.

\subsection{Privacy and Complexity Analysis}\label{sec:Priv_analysis}
The complete training process for ASGL is outlined in Algorithm~\ref{algo:ASGL} (see \textbf{App.~\ref{sec:training}}). 
In this section, we present a comprehensive privacy analysis and computational complexity analysis for ASGL.

\textbf{Privacy Accounting}. In this section, we adopt the functional perspective of Rényi Differential Privacy (RDP; see  \textbf{App.~\ref{sec:RDP}}) to analyze privacy budgets of ASGL, as summarized below:
\begin{theorem}
\label{theorem:Privacy Analysis}
Given the number of training set $N_{tr}$, number of epochs $n^{epoch}$, number of discriminators' iterations $n^{iter}$, batch size $B_d$, maximum path length $L$, and maximum path number $N$, over $T=n^{epoch}n^{iter}$ iterations, Algorithm~\ref{algo:ASGL} satisfies node-level $(\alpha, 2T\gamma)$-RDP, where $\gamma=\frac{1}{\alpha-1} \ln \left(\sum_{i=0}^{R_{N,L}} \beta_i \left(\exp{\frac{\alpha (\alpha-1)i^2 }{2 \sigma^2R_{N,L}^2}}\right)\right)$, $R_{N,L}=\frac{N^{L+1}-1}{N-1}$ and $\beta_i=\binom{R_{N,L}}{i}\binom{N_{tr}-R_{N,L}}{B_d-i}/{\binom{N_{tr}}{B_d}}$. Please refer to  \textbf{App.~\ref{app:proof1}} for the proof.
\end{theorem}

\textbf{Complexity Analysis}.
To analyze the time complexity of training ASGL (\textbf{App.~\ref{sec:training}}), we break down the major computations. The outer loop runs for $n^{\text{epoch}}$ epochs, and in each epoch, the discriminators $D^+$ and $D^-$ are trained for $n^{\text{iter}}$ iterations. Each iteration samples a batch of $B_d$ real and fake edges to update $\theta_D$, with DP cost updates incurring complexity $\mathcal{O}(B_d k \xi)$, where $\xi$ is the sampling probability and $k$ is the embedding dimension~\cite{08,17}. Thus, each epoch of $D^+$ or $D^-$ costs $\mathcal{O}(n^{\text{iter}} B_d k (1 + \xi))$.
For the generators $G^+$ and $G^-$, each iteration samples $B_g$ fake edges to update $\theta_G$, resulting in per-epoch complexity $\mathcal{O}(n^{\text{iter}} B_g k)$.
In total, ASGL's overall time complexity over $n^{\text{epoch}}$ epochs is:
$\mathcal{O}\left(2 n^{\text{epoch}} n^{\text{iter}} (B_d + B_g)(1 + \xi)k\right)$.
This complexity is linear in the number of iterations and batch size, demonstrating the scalability of ASGL for large-scale graphs.

\section{Experiments}\label{sec:experiments}
In this section, some experiments are designed to answer the following questions:
(1) How do key parameters affect the performance of ASGL (See Section~\ref{sub: Key Parameters})?
(2) How much does the privacy budget affect the performance of ASGL and other private signed graph learning models in edge sign prediction (See Section~\ref{sub: Edge Sign Prediction})?
(3) How much does the privacy budget affect the performance of ASGL and other baselines in node clustering (See Section~\ref{sub: Node Clustering})?
(4) How resilient is ASGL to defense link stealing attacks (See Section~\ref{sub:LSA})?

\begin{table}[!t]
\small
\setlength{\abovecaptionskip}{2pt}
\setlength{\belowcaptionskip}{-5pt}
\centering
  \caption{Overview of the datasets}
  \label{tab:dataset}
  \setlength{\tabcolsep}{3pt} 
  \begin{tabular}{ccccc}
    \toprule
    Datasets & Nodes &Edges &Positive Edges &Negative Edges \\
    \midrule
    Bitcoin-Alpha &3,783 &14,081 &12,769 (90.7$\%$) &1,312 (9.3$\%$)\\
    Bitcoin-OTC &5,881&21,434 &18,281 (85.3$\%$)  &3,153  (14.7$\%$)\\
    WikiRfA &11,258 &185,627 &144,451 (77.8$\%$) &41,176 (22.2$\%$)\\
    Slashdot &13,182 &36,338 &30,914 (85.1$\%$) &5,424  (14.9$\%$)\\
    Epinions &131,828 &841,372 &717,690 (85.3$\%$) &123,682  (14.7$\%$)\\
  \bottomrule
\end{tabular}
\end{table}

\subsection{Experimental Settings}
\textbf{Datasets}.
To comprehensive evaluate our ASGL method, we conduct extensive experiments on five real-world datasets, namely Bitcoin-Alpha\footnote{\label{foot:gplus}Collected in https://snap.stanford.edu/data.}, Bitcoin-OTC\footnotemark[\getrefnumber{foot:gplus}], WikiRfA\footnotemark[\getrefnumber{foot:gplus}], Slashdot\footnote{Collected in https://www.aminer.cn.} and Epinions\footnotemark[\getrefnumber{foot:gplus}]. These datasets are regarded as undirected signed graphs, with their detailed statistics summarized in Table~\ref{tab:dataset} and  \textbf{App.~\ref{app:data}}.

\textbf{Competitive Methods}. To the best of our knowledge, this work is the first to address the problem of differentially private signed graph learning while aiming to preserve model utility. Due to the absence of prior studies in this area, we construct baselines by integrating four state-of-the-art signed graph learning methods—SGCN~\cite{36}, SiGAT~\cite{38}, LSNE~\cite{37}, and SDGNN~\cite{39}—with the DPSGD mechanism. Since these models primarily leverage structural information, we further include the private graph learning method GAP~\cite{40}, using Truncated SVD-generated spectral features~\cite{36} as input to ensure a fair comparison involving node features.

\textbf{Evaluation Metrics}.
For edge sign prediction tasks, we follow the evaluation procedures in~\cite{14,38,39}. Specifically, we first generate embedding vectors for all nodes in the training set using each comparative method. Then, we train a logistic regression classifier using the concatenated embeddings of node pairs as input features. Finally, we use the trained classifier to predict edge signs in the test set for each method. Considering the class imbalance between positive and negative edges (see Table~\ref{tab:dataset}), we adopt the \textit{area under curve} (AUC) as the evaluation metric to ensure a fair comparison.

For node clustering, to fairly evaluate the clustering effect of node embeddings, we compute the average cosine distance for both positive and negative node pairs:
$\text{CD}^+=\sum_{(v_i,v_j)\in E^+} Cos(\mathbf{Z}_i,\mathbf{Z}_j)/|E^+|$ and $\text{CD}^-=\sum_{(v_n,v_m)\in E^-} Cos(\mathbf{Z}_n,\mathbf{Z}_m)/|E^-|$, where $\mathbf{Z}_i$ is the node embedding generated by each comparative method, and $Cos(\cdot)$ represents the cosine distance between node embeddings. Then we propose the \textit{symmetric separation index} (SSI) to measure the clustering degree between the embeddings of positive and negative node pairs in the test set, denoted as $\text{SSI}=1/(|\text{CD}^+-1|+|\text{CD}^-+1|)$. A higher SSI indicates better structural proximity, with positive node pairs more tightly clustered and negative pairs more clearly separated in the unified embedding space.

\textbf{Parameter Settings}. For both edge sign prediction and node clustering tasks, we set the dimensionality of all node embeddings, \( \mathbf{d}_v \) and \( \mathbf{g}_v \), to 128, following standard practice in prior work~\cite{41,14}. ASGL adopts DPSGD-based optimization, where the total number of training epochs is determined by the moments accountant (MA)~\cite{04}, which offers tighter privacy tracking across multiple iterations.
We set the iteration number $n^{iter}$ to 10 for Bitcoin-Alpha and Bitcoin-OTC, 15 for WikiRfA and Slashdot, and 20 for Epinions. Since all comparative methods are trained using DPSGD, their number of training epochs depends on the privacy budget.
As discussed in Section~\ref{sub: Key Parameters}, the maximum path number \( N \) and path length \( L \) are varied to analyze their impact on ASGL’s utility.
For privacy parameters, we follow~\cite{02,51,08} by fixing \( \delta = 10^{-5} \) and \( C = 1 \), and vary the privacy budget \( \epsilon \in \{1, 2, \dots, 6\} \) to evaluate utility under different privacy levels. To ensure fair comparison, we modify the official GitHub implementations of all baselines and adopt the best hyperparameter settings reported in their original papers. To minimize random errors, each experiment is repeated five times.

\subsection{Impact of Key Parameters}
\label{sub: Key Parameters}
In this section, 
we perform experiments on two datasets by varying the maximum number $N$ and the maximum length $L$ of paths in the BFS-trees, providing a rationale for parameter selection.

\subsubsection{The effect of the parameter $N$}
As discussed in Section~\ref{sub:Constrained BFS-trees}, the greater the number of neighbors a rooted node has, the more paths can be obtained through random walks. Therefore, the maximum number of paths $N$ also depends on the node degrees. As shown in Fig.~\ref{fig:dis_degree} (see \textbf{App.~\ref{app:NL}}), for the Bitcoin-Alpha and Slashdot datasets, most nodes in signed graphs have degrees below 3. In addition, we investigate the impact of $N$ by varying its value within $\{2,3,4,5,6\}$. As shown by the average AUC results in Table~\ref{tab:path_count}, the proposed ASGL method achieves optimal edge prediction performance at $N=3$ for Bitcoin-Alpha and $N=4$ for Slashdot. Considering both gradient sensitivity and computational efficiency, we adopt $N=3$ for subsequent experiments.

\subsubsection{The effect of the parameter $L$}
In this experiment, we evaluate the impact of the path length $L$ on the utility of ASGL by varying its value. As shown in Table~\ref{tab:path_len}, ASGL achieves the best performance on both datasets when $L=4$. This result is closely aligned with the structural characteristics of the signed graphs:
As summarized in Fig.~\ref{fig:dis_path_len} (see \textbf{App.~\ref{app:NL}}), most node pairs in these datasets exhibit maximum path lengths of 3 or 4. Therefore, in subsequent experiments, we set $L=4$, as it adequately covers the receptive field of most nodes.

\vspace{-8pt}
\begin{table}[!h]
\centering
\begin{threeparttable}
 \captionsetup{skip=1pt}  
  \caption{Summary of average AUC with different maximum path counts $N$ under $\epsilon=3$ and $L=3$. (\textbf{BOLD}: Best)}
\label{tab:path_count}
  \begin{tabular}{cccccc}
    \toprule
    Dataset& $N=2$ &$N=3$ &$N=4$ &$N=5$ &$N=6$\\

    \midrule
    \multirow{1}{*}{Bitcoin-Alpha} &0.8025 &\textbf{0.8562} &0.8557 &0.8498 &0.8553 \\
    
    \multirow{1}{*}{Slashdot}
    &0.7723 &0.8823 &\textbf{0.8888} &0.8871 &0.8881  \\
  \bottomrule
\end{tabular}
\end{threeparttable}
\end{table}

\vspace{-15pt}
\begin{table}[!h]
\centering
\begin{threeparttable}
\captionsetup{skip=1pt} 
  \caption{Summary of average AUC with different path lengths $L$ under $\epsilon=3$ and $N=3$. (\textbf{BOLD}: Best)}
\label{tab:path_len}
  \begin{tabular}{cccccc}
    \toprule
    Dataset& $L=1$ &$L=2$ &$L=4$ &$L=6$ &$L=8$\\

    \midrule
    \multirow{1}{*}{Bitcoin-Alpha} &0.7409 &0.8443 &\textbf{0.8587} &0.8545 &0.8516 \\
    
    \multirow{1}{*}{Slashdot}
    &0.7629 &0.8290 &\textbf{0.8833} &0.8809 &0.8807  \\
  \bottomrule
\end{tabular}
\end{threeparttable}
\end{table}
\vspace{-8pt}

\begin{figure*}[!h]
\setlength{\abovecaptionskip}{2pt}
\setlength{\belowcaptionskip}{-5pt}
  \centering
  \includegraphics[width=1\linewidth]{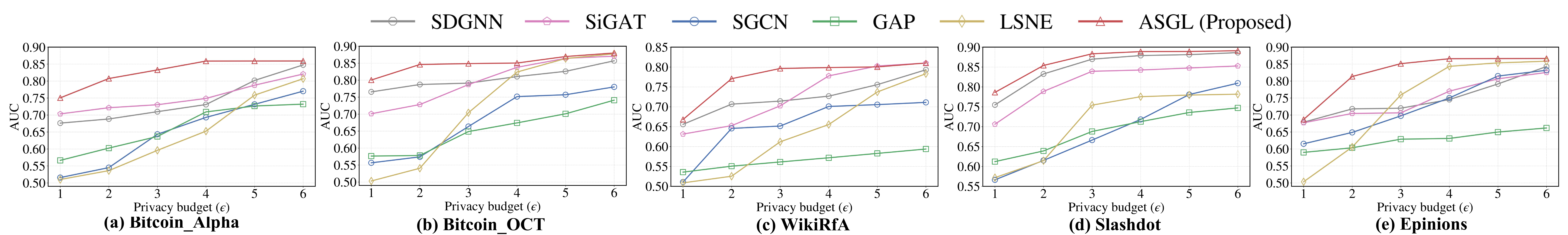}
  \caption{AUC vs. Privacy cost ($\epsilon$) of private signed graph learning methods in edge sign prediction.}
  \label{fig:edge sign prediction}
\end{figure*}

\begin{figure*}[!h]
\setlength{\abovecaptionskip}{2pt}
\setlength{\belowcaptionskip}{-5pt}
  \centering
  \includegraphics[width=1\linewidth]{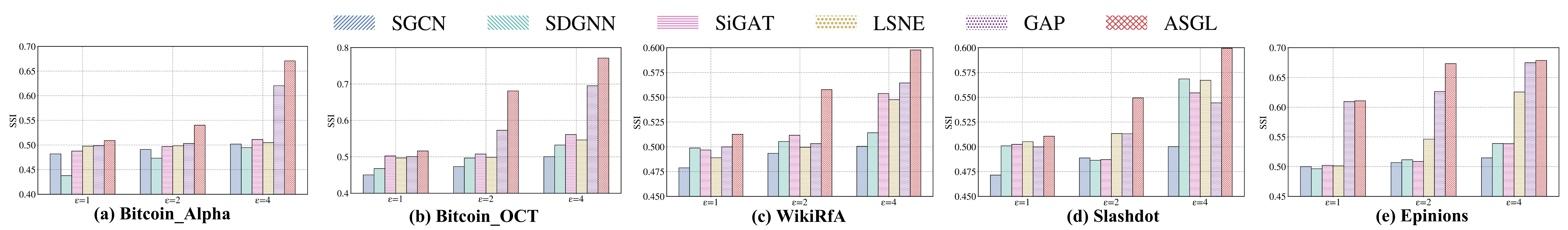}
  \caption{Symmetric separation index (SSI) vs. Privacy cost ($\epsilon$) of private signed graph learning methods in node clustering.}
  \label{fig:node clustering}
\end{figure*}

\subsection{Impact of Privacy Budget on Edge Sign Prediction}
\label{sub: Edge Sign Prediction}
To evaluate the effectiveness of different private graph learning methods on edge sign prediction, we compare their AUC scores under privacy budgets $\epsilon$ ranging from 1 to 6, as shown in Fig.~\ref{fig:edge sign prediction} and Table~\ref{tab:detailed ESP} (see  \textbf{App.~\ref{app:edge_sign_pre}}). The proposed ASGL consistently outperforms all baselines across all privacy levels and datasets, owing to its ability to generate node embeddings that preserve connectivity distributions while satisfying DP guarantees.
Although SDGNN achieves sub-optimal performance, it exhibits a noticeable gap from ASGL under limited privacy budgets (\( \epsilon < 4 \)). SiGAT, SGCN, and LSNE employ the moments accountant (MA) to mitigate excessive privacy budget consumption, yet still suffer from poor convergence and degraded utility under limited privacy budgets. GAP adopts aggregation perturbation to ensure node-level DP, but its performance is limited due to noisy neighborhood information, hindering its ability to capture structural information for edge prediction tasks.

\subsection{Impact of Privacy Budget on Node Cluster}
\label{sub: Node Clustering}
To further examine the capability of ASGL in preserving signed node proximity, we conduct a fair comparison across multiple private graph learning methods using the SSI metric. As shown in Fig.~\ref{fig:node clustering} and Table~\ref{tab:detailed nc}  (see  \textbf{App.~\ref{app:node_cluster}}), ASGL consistently outperforms all baselines across different datasets and privacy budgets, demonstrating that ASGL is capable of generating node embeddings that effectively preserve signed node proximity. Notably, GAP achieves the second-best clustering performance on most datasets (excluding Slashdot), benefiting from its ability to leverage node features for clustering nodes. Nevertheless, to guarantee node-level DP, GAP needs to repeatedly query sensitive graph information in every training iteration, resulting in significantly higher privacy costs.

\subsection{Resilience Against Link Stealing Attack}
\label{sub:LSA}
To assess the effectiveness of ASGL in preserving the privacy of edge information, we perform link stealing attacks (LSA) across all datasets and compare the resilience of all methods to such attacks in edge sign prediction tasks. The LSA setup is detailed in  \textbf{App.~\ref{sec:setup_LSA}}. Attack performance is measured by the AUC score, averaged over five independent runs.
Table~\ref{tab:LSA} summarizes the effectiveness of LSA on various trained target models and datasets. It can be observed that as the privacy budget $\epsilon$ increases, the average AUC of LSA consistently improves, indicating the reduced privacy protection of target models and an increased success rate of the attack. Overall, the average AUC of the attack is close to 0.50 in most cases, indicating the unsuccessful edge inference and the robustness of DP against such an attack. When $\epsilon = 3$, ASGL demonstrates stronger resistance to LSA across most datasets, with AUC values consistently below 0.57. This suggests that ASGL offers defense performance comparable to other differentially private graph learning methods. 

 \vspace{-6pt}
\begin{table}[!h]  
\small
\centering
\setlength{\tabcolsep}{3pt}  
\begin{threeparttable}
\captionsetup{skip=1pt} 
  \caption{The average AUC of LSA on different comparisons and datasets. (\textbf{BOLD}: Best resilience against LSA)}
  \label{tab:LSA}
  \begin{tabular}{cccccccc}
    \toprule
    $\epsilon$ & Dataset &GAP &SGCN &LSNE &SiGAT &SDGNN &ASGL\\

    \midrule
    \multirow{5}{*}{1}
    &Bitcoin-Alpha &0.5072 &0.7091 &0.5079 &0.5145 &0.5404 &\textbf{0.5053}\\
    &Bitcoin-OTC &\textbf{0.5081} &0.7118 &0.5119 &0.5409 &0.5660 &0.5466\\
    &Slashdot &0.5538 &0.8232 &0.5551 &0.5609 &0.5460 &\textbf{0.5325}\\
    &WikiRfA &\textbf{0.5148} &0.5424 &0.5427 &0.5293 &0.5470 &0.5302\\
    &Epinions &0.7877 &0.6329 &0.5114 &0.5129 &0.5188 &\textbf{0.5092}\\

    \midrule
    \multirow{5}{*}{3}
    &Bitcoin-Alpha &0.5547 &0.7514 &0.5533 &0.5542 &0.5598 &\textbf{0.5430}\\
    &Bitcoin-OTC &0.5655 &0.7273 &0.5684 &0.5734 &0.5765 &\textbf{0.5612}\\
    &Slashdot &0.5742 &0.8394 &0.6267 &0.5730 &0.6464 &\textbf{0.5634}\\
    &WikiRfA &\textbf{0.5276} &0.5466 &0.5542 &0.5696 &0.5772 &0.5624\\
    &Epinions &0.7981 &0.6456 &0.5588 &0.5629 &0.5665 &\textbf{0.5542}\\
  \bottomrule
\end{tabular}
\end{threeparttable}
\end{table}
 \vspace{-8pt}

\section{Related Work}\label{sec:related_work}
\textbf{Signed graph learning.}
In recent years, deep learning approaches have been increasingly adopted for signed graph learning. For example, SiNE~\cite{47} extracts signed structural information based on balance theory and designs an objective function to learn signed node proximity. Furthermore, the GNN model~\cite{36} and its variants~\cite{38,39} are used to learn signed relationships between nodes in multi-hop neighborhoods. However, these GNNs-based methods depend on the message-passing mechanism, which is sensitive to noisy interactions between nodes~\cite{49}. 
To address this issue,  Lee \emph{et al}.~\cite{14} extends the adversarial framework to signed graphs by generating both positive and negative node embeddings.
Still, these signed graph learning models are vulnerable to user-linkage attacks.

\textbf{Private graph learning.}
Recent works have increasingly focused on developing DP methods to address privacy leakage in GNNs. For instance, Daigavane \emph{et al}.~\cite{33} propose a DP-GNN method based on gradient perturbation. However, this method fails to balance utility and privacy due to excessive noise. 
Furthermore, GAP~\cite{40} and DPRA~\cite{50} are proposed to ensure the privacy of sensitive node embeddings by perturbing node aggregations.
Despite their success in node classification, the private node information is repeatedly queried in the training process of GAP, which consumes more privacy budgets to implement DPSGD. DPRA is not well-suited for signed graph embedding learning, as its edge perturbation strategy introduces cascading errors under balance theory.

\section{Conclusion}\label{sec:conclusion}
In this paper, we propose ASGL that achieves strong model utility while providing node-level DP guarantees. To address the cascading error and gradient polarity reversals from edge sign flips, ASGL separately processes positive and negative subgraphs within a shared embedding space using a DPSGD-based adversarial mechanism to learn high-quality node embeddings.
To further reduce gradient sensitivity, we introduce a constrained BFS-tree strategy that limits node receptive fields and enables gradient decoupling. This effectively reduces the required noise scale and enhances model performance.
Extensive experiments demonstrate that ASGL achieves a favorable privacy-utility trade-off. Our future work is to extend the ASGL framework by considering edge directions and weights.

\begin{acks}
This work was supported by the National Natural Science Foundation of China (Grant No:  62372122 and 92270123),  and the Research Grants Council (Grant No: 15208923, 25207224, and 15207725), Hong Kong SAR, China.
\end{acks}

\bibliographystyle{ACM-Reference-Format}
\bibliography{lib}

\appendix
\section{Adversarial Learning on Graph}
\label{app:AL}
The adversarial learning model for graph embedding~\cite{03} is illustrated as follows. Let $\mathcal{N}(v_r)$ be the node set directly connected to $v_r$. We denote the underlying true connectivity distribution of node $v_r$ as the conditional probability $p(v|v_r)$, which captures the preference of $v_r$ to connect with other nodes $v \in V$. In other words, the neighbor set $\mathcal{N}(v_r)$ can be interpreted as a set of observed nodes drawn from $p(v|v_r)$. The adversarial learning for the graph $\mathcal{G}$ is conducted by the following two modules:

\textbf{Generator $G$}: Through optimizing the generator parameters $\theta_G$, this module aims to approximate the underlying true connectivity distribution and generate (or select) the most likely nodes $v\in V$ that are relevant to $v_r$. Specifically, the \textbf{\textit{fake}}\footnote{The term “Fake” indicates that although a node $v$ selected by the generator is relevant to $v_r$, there is no actual edge between them.} (i.e., estimated) connectivity distribution of node $v_r$ is calculated as:
\begin{equation}
\label{eq:01}
p^\prime(v | v_r)= G\left(v | v_r;\theta_G\right)=\frac{\exp \left(\mathbf{g}_v^{\top} \mathbf{g}_{v_r}\right)}{\sum_{v \neq v_r} \exp \left(\mathbf{g}_v^{\top} \mathbf{g}_{v_r}\right)},
\end{equation}
where $\mathbf{g}_v, \mathbf{g}_{v_r} \in \mathbb{R}^k$ are the $k$-dimensional vectors of nodes $v$ and $v_r$, respectively, and $\theta_G$ is the union of all $\mathbf{g}_v$'s. To update $\theta_G$ in each iteration, a set of node pairs $(v,v_r)$, not necessarily directly connected, is sampled according to $p^\prime(v | v_r)$. The key purpose of generator $G$ is to deceive the discriminator $D$, and thus its loss function $L_G$ is determined as follows:
\begin{equation}
\begin{aligned}
L_G=\min\sum_{r=1}^{|V|} \left.\mathbb{E}_{v \sim G\left(\cdot \mid v_r ; \theta_G\right)}\left[\log \left(1-D\left(v_r, v \mid \theta_D\right)\right)\right]\right.,
\end{aligned}
\end{equation}
where the discriminant function $D(\cdot)$ estimates the probability that a given node pairs $(v,v_r)$ are considered \textbf{\textit{real}}, i.e., directly connected. 

\textbf{Discriminator $D$}: This module tries to distinguish between real node pairs and fake node pairs synthesized by the generator $G$. Accordingly, the discriminator estimates the probability that an edge exists between $v_r$ and $v$, denoted as:
\begin{equation}
\label{eq:02}
D(v_r,v|\theta_D) = \sigma(\mathbf{d}_v^\top  \mathbf{d}_{v_r}) = \frac{1}{1+\exp({-\mathbf{d}_v^\top  \mathbf{d}_{v_r})}},
\end{equation}
where $\mathbf{d}_v, \mathbf{d}_{v_r} \in \mathbb{R}^k$ are the $k$-dimensional vectors corresponding to the $v$-th and $v_r$-th rows of discriminator parameters $\theta_D$, respectively. $\sigma(\cdot)$ represents the sigmoid function of the inner product of these two vectors. Given the sets of real and fake node pairs, the loss function of $D$ can be derived as:
\begin{equation}
\begin{aligned}
L_D=&\max\sum_{r=1}^{|V|}\left(\mathbb{E}_{v \sim p\left(\cdot \mid v_r\right)}\left[\log D\left(v, v_r \mid \theta_D\right)\right]\right. \\
& \left.+\mathbb{E}_{v \sim G\left(\cdot \mid v_r ; \theta_G\right)}\left[\log \left(1-D\left(v_r, v \mid \theta_D\right)\right)\right]\right).
\end{aligned}
\end{equation}

In summary, the generator $G$ and discriminator $D$ operate as two adversarial components: the generator $G$ aims to fit the true connectivity distribution $p(v | v_r)$, generating candidate nodes $v$ that resemble the real neighbors of $v_r$ to deceive the discriminator $D$. In contrast, the discriminator $D$ seeks to distinguish whether a given node is a true neighbor of $v_r$ or one generated by $G$. Formally, $D$ and $G$ are engaged in a two-player minimax game with the following loss function:
\begin{equation}
\label{eq:03}
\begin{aligned}
\min _{\theta_G} & \max _{\theta_D} L(G, D)=\sum_{r=1}^{|V|}\left(\mathbb{E}_{v \sim p\left(\cdot \mid v_r\right)}\left[\log D\left(v, v_r \mid \theta_D\right)\right]\right. \\
& \left.+\mathbb{E}_{v \sim G\left(\cdot \mid v_r ; \theta_G\right)}\left[\log \left(1-D\left(v_r, v \mid \theta_D\right)\right)\right]\right).
\end{aligned}
\end{equation}

Based on Eq.~\eqref{eq:03}, the parameters $\theta_D$ and $\theta_G$ are updated by alternately maximizing and minimizing the loss function $L(G, D)$. Competition between $G$ and $D$ results in mutual improvement until $G$ becomes indistinguishable from the true connectivity distribution.

\section{Notation Introduction}\label{app:notations}
The frequently used notations are summarized in~\autoref{tab:01}.
\begin{table}[!h]
\small
\centering
  \captionsetup{skip=1pt} 
\caption{Notation Summary}
\label{tab:01}
\begin{tabular}{c|l}
\hline
\textbf{Symbol} & \textbf{Description} \\ 
\hline
$\mathcal{G},\mathcal{G^+},\mathcal{G^-}$ & Signed graph, positive subgraph, negative subgraph \\
$V, E^+, E^-$ & Node set, negative and positive edge sets\\
$\mathcal{N}(v_r)$ & Neighbor node set of node $v_r$ \\
$\theta_D$ & Shared parameters of discriminators $D^+$ and $D^-$\\
$\theta_G$ & Shared parameters of generators $G^+$ and $G^-$\\
$\mathbf{d}_{v_r}$ & Node embedding for node $v_r$ of Discriminators\\
$\mathbf{g}_{v_r}$ & Node embedding for node $v_r$ of Generators\\
$N, L$ & Maximum number and length of generated path \\
$\epsilon, \delta$ & Privacy parameters  \\
$\mathcal{N}(0,\sigma^2)$ & Gaussian distribution with standard deviation $\sigma^2$\\
$P_{rt}$ & A path from rooted node $v_r$ to target node $v_t$ \\
$T^+_{v_r}, T^-_{v_r}$ & Positive and negative BFS-trees rooted from $v_r$ \\
$p_{\text{true}}^{+}(\cdot | v_r)$ & Positive connectivity distributions of $(v_r,v)\in E^+$\\
$p_{\text{true}}^{-}(\cdot | v_r)$ & Negative connectivity distributions of $(v_r,v)\in E^-$\\
$p^+_{T^+_{v_r}}(v | v_r)$ & Positive relevance probability between $v_r$ and $v$\\
$p^-_{T^-_{v_r}}(v | v_r)$ & Negative relevance probability between $v_r$ and $v$\\
\hline
\end{tabular}
\end{table}

\section{Rényi Differential Privacy}
\label{sec:RDP}
Since standard DP can be overly strict for deep learning, we follow prior work~\cite{30,31} and adopt an alternative definition—Rényi Differential Privacy (RDP)~\cite{07}. RDP offers tighter and more efficient composition bounds, enabling more accurate estimation of cumulative privacy cost over multiple queries on graphs.

\begin{definition}[Rényi Differential Privacy~\cite{07}] The Rényi divergence quantifies the similarity between output distributions of a mechanism and is defined as:
\begin{equation}
\small
D_\alpha(P \| Q)=\frac{1}{\alpha-1} \log \left(\sum_x P(x)^\alpha Q(x)^{1-\alpha}\right),
\end{equation}
where $P(x)$ and $Q(x)$ are probability distributions over the output space. $\alpha>1$ denotes the order of the divergence, and its choice allows for different levels of sensitivity to the output distribution.
Accordingly, an algorithm $\mathcal{M}$ satisfies $(\alpha,\epsilon)$-RDP if, for any two adjacent graphs $\mathcal{G}$ and $\mathcal{G}^\prime$, the following condition holds
$D_\alpha\left(\mathcal{M}(\mathcal{G}) \| \mathcal{M}\left(\mathcal{G}^{\prime}\right)\right) \leq \epsilon$.
\end{definition}

Since RDP is an extension of DP, it can be converted into ($\epsilon$,$\delta$)-DP based on Proposition 3 in~\cite{07}, as outlined below.

\begin{lemma}[Conversion from RDP to DP~\cite{07}] 
If a mechanism $\mathcal{M}$ satisfies $(\alpha,\epsilon)$-RDP, it also satisfies $(\epsilon+\log (1 / \delta) / (\alpha-1), \delta)$-DP for any $\delta \in (0,1)$.
\label{lemma:01}
\end{lemma}

\section{Gaussian Mechanism}
Let $f$ be a function that maps a graph $\mathcal{G}$ to $k$-dimensional node vectors $\mathbf{Z}\in\mathbb{R}^{|V|\times k}$. To ensure the RDP guarantees of $f$, it is common to inject Gaussian noise into its output~\cite{07}. The noise scale depends on the sensitivity of $f$, defined as $\Delta_f=\max _{\mathcal{G}, \mathcal{G}^{\prime}}\left\|f(\mathcal{G})-f\left(\mathcal{G}^{\prime}\right)\right\|_2$. Specifically, the privatized mechanism is defined as $\mathcal{M}(\mathcal{G})=f(\mathcal{G})+\mathcal{N}(0,\sigma^2 \mathbf{I})$, where $\mathcal{N}(0,\sigma^2 \mathbf{I})$ is the Gaussian distribution with zero mean and standard deviation $\sigma^2$. This results in an $(\alpha,\epsilon)$-RDP mechanism $\mathcal{M}$ for all $\alpha>1$ with $\epsilon=\alpha\Delta_f^2/2\sigma^2$.

\begin{algorithm}[!t]
\small
\caption{SAMPLE-SUBGRAPHS by Constrained BFS-trees}
\label{algo:SAMPLE-SUBGRAPHS}
\KwIn{Graph $\mathcal{G} = \{\mathcal{G}^+,\mathcal{G}^-\}$; The training set of nodes $V_{tr}$; The maximum path length $L$; The maximum path number $N$.}
\KwOut{The training set of subgraphs $S_{tr}$.}
\For{$v_r \in V_{tr}$}
{
Construct BFS-trees $T^+_{v_r}$ (or $T^-_{v_r}$) rooted from the node $v_r$ on $\mathcal{G}^+$ (or $\mathcal{G}^-$); \\
\For{$n=0; n<N$}
    {
    Based on the positive and negative relevance probability in Eqs.~\eqref{eq:11} and~\eqref{eq:13}, conduct the random walk at $T^+_{v_r}$ (or $T^-_{v_r}$) to form a path $P_{rt}^{(n)+}$ (or $P_{rt}^{(n)-}$) of length $L$;\\
    Add all nodes $v$ (excluding those in $\mathcal{N}(v_r)$) along the path $P_{rt}^{(n)+}$ (or $P_{rt}^{(n)-}$) as a fake edge $(v_r, v)$ to the corresponding subgraph set $S_{tr}^+$ (or $S_{tr}^-$);\\
    Drop $P_{rt}^{(n)+}$ (or $P_{rt}^{(n)-}$) from $T^+_{v_r}$ (or $T^-_{v_r}$).
    }
}
\textbf{Return} $S_{tr} = \{S_{tr}^+,S_{tr}^-\}$;
\end{algorithm}

\section{BFS-tree Strategy}
\label{sec:BFS}
\begin{figure}[!h]
  \centering
  \includegraphics[width=0.9\linewidth]{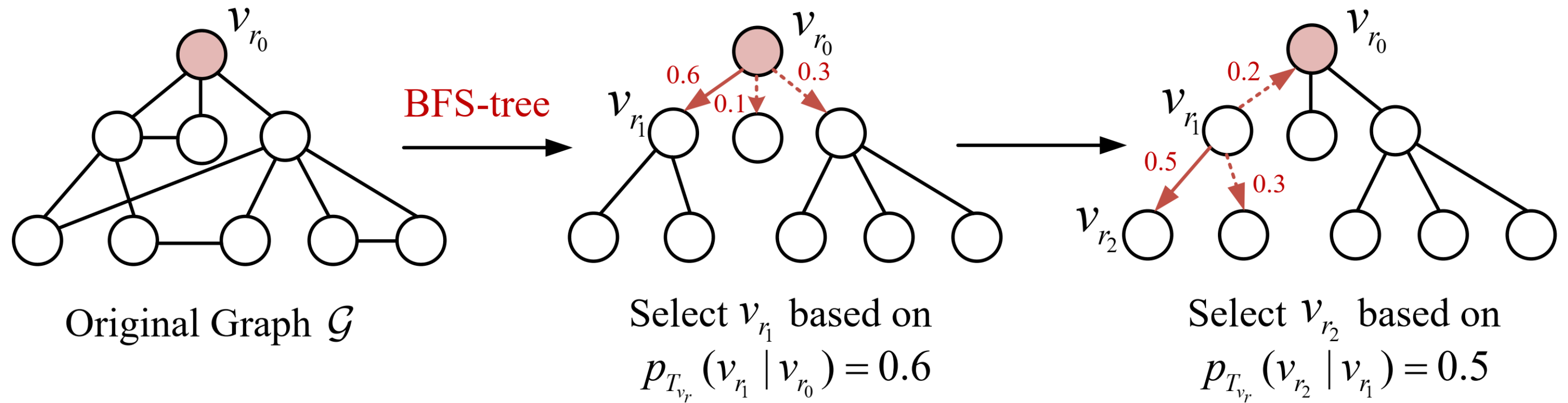}
  \captionsetup{skip=1pt}  
  \caption{Random-walk-based edge generation for generator $G^+$ or $G^-$. Red digits denote the transition probabilities (Eqs.~\eqref{eq:11} and~\eqref{eq:13}), and red arrows indicate the walk directions.}
  \label{fig:random walk}
\end{figure}

Fig.~\ref{fig:random walk} provides an illustrative example of the BFS-tree strategy: Let $v_{r_0}$ be the rooted node. We first compute the transition probabilities between $v_{r_0}$ and its neighbors $\mathcal{N}({v_{r_0}})$. The next node $v_{r_1}$ is then sampled as the first step of the walk, in proportion to these transition probabilities. Similarly, the next node $v_{r_2}$ is selected based on the transition probabilities between $v_{r_1}$ and its neighbors $\mathcal{N}({v_{r_1}})$. The random walk continues until it reaches the terminal node $v_{r_n}$, and unconnected node pairs $(v_{r_0}, v_{r_k})^+$ for $k = 2, 3, \ldots, n$  are regarded as fake positive edges.

\section{Details of Algorithm}
\label{sec:algorithm in app}
\subsection{The Parameter Update of Generators}
Given fake positive/negative edges $(v_{r}, v_{t})$ from ${G}^+/{G}^-$, the gradient of joint loss function (Eq.~\eqref{eq: loss_t}) with respect to $\theta_G$ is derived via the policy gradient~\cite{03}:  
\begin{equation}
\label{eq:L_G}
\small
 \nabla L_{G} =\left\{\begin{array}{l}
\sum_{r=1}^{|V^+|}[\nabla_{\theta_G} \log G^{+}\left(v_t | v_r; \theta_G\right) \log \left(1-D^{+}\left(v_t, v_r\right)\right)], \\
\text {if }\left(v_r, v_t\right) \text{ is a fake positive edge}; \\
\sum_{r=1}^{|V^-|}\nabla_{\theta_G} \log G^{-}\left(v_t | v_r; \theta_G\right) \log \left(1-D^{-}\left(v_t, v_r\right)\right), \\
\text {if }\left(v_r, v_t\right)\text{ is a fake negative edge}.
\end{array}\right.
\end{equation}

\subsection{SAMPLE-SUBGRAPHS by Constrained BFS-trees}
\label{sec:BFS-TREE}
As shown in Algorithm~\ref{algo:SAMPLE-SUBGRAPHS}, during the random walk on the BFS tree \( T^+_{v_r} \) or \( T^-_{v_r} \) rooted at \( v_r \in V_{tr} \), we generate multiple unique paths while constraining their number and length by parameters \( N \) and \( L \), respectively. This process yields a training subgraph set \( S_{tr} \) composed of constrained paths.

\begin{algorithm}[!h]
\small
\caption{ASGL Algorithm}
\label{algo:ASGL}
\KwIn{Graph $\mathcal{G}$; Training set of nodes $V_{tr}$; Maximum path length $L$; Maximum path number $N$; Batch-size $B_d$ and $B_g$ of sampled edges in discrininators and generators; Number of epochs $n^{epoch}$; Number of iterations for generators and discriminators per epoch $n^{iter}$; Privacy parameters $\delta$, $\epsilon$, $\sigma$.}
\KwOut{Privacy-preserving node embedding $\mathbf{g}_v \in \theta_G$ for downstream tasks.}
According to edge signs, divide $\mathcal{G}$ into $\mathcal{G}^+$ and $\mathcal{G}^-$; \\
Generate the training subgraph set $S_{tr} = \{S_{tr}^+, S_{tr}^-\}$ based on $\text{SAMPLE-SUBGRAPHS}(\mathcal{G},V_{tr}, N, L)$ in Algorithm~\ref{algo:SAMPLE-SUBGRAPHS}; \\
\For{$v_r \in V_{tr}$}
{
Sample all real positive edges $(v_r,v_t)^+$ from $\mathcal{G}^+$; \\
Sample all fake positive edges $(v_r,v_t^{\prime})^+$ from $S_{tr}^+$; \\
Sample all real negative edges $(v_r,v_t)^-$ from $\mathcal{G}^-$; \\
Sample all fake negative edges $(v_r,v_t^{\prime})^-$ from $S_{tr}^-$; \\
$E^+_D.add((v_r,v_t)^+,(v_r,v_t^{\prime})^+)$, $E^+_G.add((v_r,v_t^{\prime})^+)$, \\$E^-_D.add((v_r,v_t)^-,(v_r,v_t^{\prime})^-)$, $E^-_G.add((v_r,v_t^{\prime})^-)$; \\
}
\For{$epoch=0; epoch<n^{epoch}$}
{   
    \textit{Train the discriminator $D^+$}:\\
    \For{$iter=0; iter<n^{iter}$}
    {
        Sample $B_d$ real and fake positive edges from $E^+_D$;\\
        Update $\theta_D$ via Eqs.~\eqref{eq:D+} and~\eqref{eq: LD_all}, and achieve gradient perturbation via Eq.~\eqref{eq:L_D_NEW};\\
        Calculate privacy spent $\hat{\delta}$ given the target $\epsilon$;\\
        Stop optimization if $\hat{\delta} \geq \delta$.
    }
    \textit{Train the generator $G^+$}: \\
    \For{$iter=0; iter<n^{iter}$}
    {
        Subsample $B_g$ fake positive edges from $E^+_G$; \\
        Update $\theta_G$ via Eqs.~\eqref{eq: FPCP} and~\eqref{eq:L_G}. 
    }
    \textit{Train the discriminator $D^-$}: \\
    \For{$iter=0; iter<n^{iter}$}
    {
        Subsample $B_d$ real and fake negative edges from $E^-_D$;\\
        Update $\theta_D$ via Eqs.~\eqref{eq:D-} and~\eqref{eq: LD_all}, and achieve gradient perturbation via Eq.~\eqref{eq:L_D_NEW};\\
        Calculate privacy spent $\hat{\delta}$ given the target $\epsilon$;\\
        Stop optimization if $\hat{\delta} \geq \delta$. 
    }
    \textit{Train the generator $G^-$}: \\
    \For{$iter=0; iter<n^{iter}$}
    {
        Subsample $B_g$ fake negative edges from $E^-_G$;\\
        Update $\theta_G$ via Eqs.~\eqref{eq: FNCP} and~\eqref{eq:L_G}.
    }
}
\textbf{Return} privacy-preserving node embedding $\mathbf{g}_v \in \theta_G$\;
\end{algorithm}

\subsection{The training of ASGL}
\label{sec:training}
The training process of ASGL is outlined in Algorithm~\ref{algo:ASGL} and consists of the following main steps:

(1) \textbf{Signed graph decomposition and subgraph sampling}: Given an input signed graph $\mathcal{G}$, we first divide it into a positive subgraph $\mathcal{G}^+$ and a negative subgraph
$\mathcal{G}^-$ based on edge signs. Then, for each node $v_r \in V_{tr}$, constrained BFS trees are constructed from $\mathcal{G}^+$ and $\mathcal{G}^-$, respectively, to generate a set of training subgraphs $S_{tr} = \{S_{tr}^+, S_{tr}^-\}$ by limiting the maximum number of paths $N$ and the maximum path length $L$. These subgraphs are used to sample fake edges for adversarial training.

(2) \textbf{Edge sampling for adversarial learning}: For each node $v_r$, we sample real edges from $\mathcal{G}^+$ and $\mathcal{G}^-$, and fake edges from $S_{tr}^+$ and $S_{tr}^-$. These edges are organized into four sets: 
\begin{itemize}[leftmargin=5mm]
\item $E_D^+$: real and fake positive edges for training $D^+$.
\item $E_G^+$: fake positive edges for training $G^+$.
\item $E_D^-$: real and fake negative edges for training $D^-$.
\item $E_G^-$: fake negative edges for training $G^-$.
\end{itemize}

(3) \textbf{Adversarial training with DPSGD}: The training is performed over $n^{epoch}$ epochs. In each epoch:
\begin{itemize}[leftmargin=5mm]
\item Discriminator training: For each discriminator $D^+$ and $D^-$, we perform $n^{iter}$ iterations. In each iteration, a batch of $B_d$ real and fake edges is sampled. The discriminator parameters $\theta_D$ are updated using gradient descent with noise addition according to the DPSGD mechanism (Eq.~\eqref{eq:L_D_NEW}), ensuring node-level DP. The privacy budget $\hat{\delta}$ is tracked, and training stops early if $\hat{\delta}>\delta$.
\item Generator training: Each generator $G^+$ and $G^-$ is trained for $n^{iter}$ iterations. In each iteration, a batch of $B_g$ fake edges is sampled, and the generator parameters $\theta_G$ are updated by maximizing the generator objective (Eq.~\eqref{eq:L_G}).
\end{itemize}

(4)  \textbf{Embedding output for downstream tasks}: After all epochs, the generator parameters $\theta_G$ encode the privacy-preserving node embeddings $\mathbf{g}_v \in \theta_G$, which are used for downstream tasks such as edge sign prediction and node clustering.

\section{Details of Lemma}
\label{app:lemma}
The following lemmas are used for proving Theorem~\ref{theorem:L_D_NEW}:
\begin{lemma}[Receptive field of a node]
\label{lemma: Receptive field}
As shown in Fig.~\ref{fig:receptive field}(b), we define the \textit{\textbf{receptive field}} of a node as the region (i.e., the set of nodes) over which it can exert influence. Accordingly, for a \textit{\textbf{subgraph}} constructed from paths sampled on constrained BFS-trees (Fig.~\ref{fig:receptive field}(b)), the maximum receptive field size of $v_r$ is given by $R_{N,L}=\sum_{l=0}^L N^l=\frac{N^{L+1}-1}{N-1}\leq B$.
\end{lemma}

\begin{lemma}
\label{lemma: S_v}
Let $S_{tr}$ denote the training set of subgraphs constructed from constrained BFS-tree paths,
and $S(v) \subset S_{tr}$ denote the subgraph subset that contains the node $v$. Since $R_{N,L}$ represents the upper bound on the number of occurrences of any node in $S_{tr}$, it follows that $ |S(v)| \leq R_{N,L}$. The proof of Lemma~\ref{lemma: S_v} is illustrated in  \textbf{App.~\ref{app:proof2}}.
\end{lemma}

\section{Proof of Lemma~\ref{lemma: S_v}}
\label{app:proof2}
\textit{Proof}. We proceed by induction~\cite{13} on the path length $L$ of the BFS-tree.

\textbf{Base case}: When $L=0$, each sampled subgraph $S(v)$ contains exactly the training node $v\in V_{tr}$ itself. Thus, every node appears in one subgraph, trivially satisfying the bound $|S(v)|=R_{N,0}=1$.

\textbf{Inductive hypothesis}: Assume that for some fixed $L\geq0$, any $v\in V_{tr}$ appears in at most $R_{N,L}$ subgraphs constructed from constrained BFS-tree paths. Let $S^L(v)$ denote a subgraph set with $L$ path length. Thus, the hypothesis is $|S^L(v)|\leq R_{N,L}$ for any $v$.

\textbf{Inductive step}: We further show that the above hypothesis also holds for $L+1$ path length: Let $T_{u^\prime}$ represent the $L$-length BFS-tree rooted at $u^\prime$. If $T_{u^\prime}\in S^{L+1}(v)$, there must exit node $u$ such that $u \in T_{u^\prime}$ and $T_{u}\in S^{L}(v)$. According to the setting of Algorithm~\ref{algo:SAMPLE-SUBGRAPHS}, the number of such nodes $u$ is at most $N$. By the hypothesis, there are at most $R_{N,L}-1$ such $u^\prime \neq v$ such that $T_{u^\prime}\in S^{L+1}(v)$. Based on these upper bounds, we can derive the upper bound matching the inductive hypothesis for $L+1$: 
\begin{equation}
\left|S^{L+1}(v)\right| \leq N \cdot(R_{N,L}-1)+1=\frac{N^{L+2}-1}{N-1}=R_{N,L+1} .
\end{equation}

By induction, the Lemma~\ref{lemma: S_v} holds for all $L\geq0$.

\section{Proof of Theorem~\ref{theorem:Privacy Analysis}}
\label{app:proof1}

The following lemmas are used for proving Theorem~\ref{theorem:Privacy Analysis}:

\begin{lemma}[Adaptation of Lemma 5 from~\cite{34}]
\label{lemma:RDP_GUASSIAN}
Let $\mathcal{N}(\mu,\sigma^2)$ represent the Gaussian distribution with mean $\mu$ and standard deviation $\sigma^2$, it holds that:
\begin{equation}
\mathcal{D}_\alpha\left(\mathcal{N}\left(\mu, \sigma^2\right) \| \mathcal{N}\left(0, \sigma^2\right)\right)=\frac{\alpha \mu^2}{2 \sigma^2}
\end{equation}
\end{lemma}

\begin{lemma}[Adaptation of Lemma 25 from~\cite{33}]
\label{lemma:05}
Assume $\mu_0,...,\mu_n$ and $\eta_0,...,\eta_n$ are probability distributions over some domain $Z$ such that their Rényi divergences satisfy: $\mathcal{D}_\alpha(\mu_0||\eta_0)\leq\epsilon_0,...,\mathcal{D}_\alpha(\mu_n||\eta_n)\leq\epsilon_n$ for some given $\epsilon_0,...,\epsilon_n$. 
Let $\rho$ be a probability distribution over $\{0,...,n\}$. Denoted by $\mu_\rho$ ($\eta_\rho$, respectively) the probability distribution on $Z$ obtained by sampling $i$ from $\rho$ and then randomly sampling from $\mu_i$ and $\eta_i$, we have:
\begin{equation}
\mathcal{D}_\alpha\left(\mu_\rho \| \eta_\rho\right) \leq \ln \mathbb{E}_{i \sim \rho}\left[e^{\varepsilon_i(\alpha-1)}\right]=\frac{1}{\alpha-1} \ln \sum_{i=0}^n \rho_i e^{\varepsilon_i(\alpha-1)}
\end{equation}
\end{lemma}

\noindent \textit{Proof of Theorem~\ref{theorem:Privacy Analysis}}. 
Consider any minibatch $\mathcal{B}_t$ randomly sampled from the training subgraph set $S_{tr}$ of Algorithm~\ref{algo:ASGL} at iteration $t$. For a subset $S(v^*) \subset S_{tr}$ containing node $v^*$, its size is bounded by $R_{N,L}$ (Lemma~\ref{lemma: S_v}). Define the random variable $\beta$ as $|S(v^*)\cap\mathcal{B}_t|$, and its distribution follows the hypergeometric distribution $\mathrm {Hypergeometric}(|S_{tr}|,R_{N,L},|\mathcal{B}_t|)$~\cite{32}:
\begin{equation}
\label{eq:beta}
\beta_i=P[\beta=i]\xlongequal[|S_{tr}|=N_{tr}]{|\mathcal{B}_t|=B_{d}}\frac{\binom{R_{N,L}}{i}\binom{N_{tr}-R_{N,L}}{B_{d}-i}}{\binom{N_{tr}}{B_{d}}}.
\end{equation}

Next, consider the training of the discriminators (Lines 12–18 and 24–30 in Algorithm~\ref{algo:ASGL}). Let $\mathcal{G}$ and $\mathcal{G}^\prime$ be two adjacent graphs differing only in the presence of node $v^*$ and its associated signed edges. Based on the gradient perturbation applied in Lines 15 and 27 of Algorithm~\ref{algo:ASGL}, we have:
\begin{equation}
\small
\begin{aligned}
\tilde{{g}}_t & ={g}_t+\mathcal{N}\left(0, \sigma^2 \Delta_{{g}}^2 \mathbf{I}\right)= \sum_{v \in \mathcal{B}_{t}}\text{Clip}_C(\frac{\partial L_D}{\partial \mathbf{d}_v})+\mathcal{N}\left(0, \sigma^2 \Delta_{{g}}^2 \mathbf{I}\right)\\
\tilde{{g}}^{\prime}_t & ={g} ^ { \prime }_t+\mathcal{N}\left(0, \sigma^2 \Delta_{{g}}^2 \mathbf{I}\right)=\sum_{v^\prime \in \mathcal{B}_{tr}^\prime}\text{Clip}_C(\frac{\partial L_D}{\partial \mathbf{d}_{v^\prime}})+\mathcal{N}\left(0, \sigma^2 \Delta_{{g}}^2 \mathbf{I}\right),
\end{aligned}
\end{equation}
where $\Delta_{{g}}=R_{N,L}C=\frac{N^{L+1}-1}{N-1}C$ (Theorem~\ref{theorem:L_D_NEW}). $\tilde{{g}}_t$ and $\tilde{{g}}^{\prime}_t$ denote the noisy gradients of $\mathcal{G}$ and $\mathcal{G}^\prime$, respectively. When $\beta=i$, their Rényi divergences can be upper bounded as:
\begin{equation}
\small
\begin{aligned}
\mathcal{D}_\alpha & \left(\tilde{{g}}_{t, i} \| \tilde{{g}}_{t, i}^{\prime}\right) \\
& =\mathcal{D}_\alpha\left({g}_{t, i}+\mathcal{N}\left(0, \sigma^2 \Delta_{{g}}^2 \mathbf{I}\right) \| {g}_{t, i}^{\prime}+\mathcal{N}\left(0, \sigma^2 \Delta_{{g}}^2 \mathbf{I}\right)\right) \\
& =\mathcal{D}_\alpha\left(\mathcal{N}\left({{g}}_{t, i}, \sigma^2 \Delta_{{g}}^2 \mathbf{I}\right) \| \mathcal{N}\left({{g}^{\prime}}{ }_{t, i}, \sigma^2 \Delta_{{g}}^2 \mathbf{I}\right)\right) \\
& \stackrel{(a)}{=} \mathcal{D}_\alpha\left(\mathcal{N}\left(\left({{g}}_{t, i}-{{g}}_{t, i}^{\prime}\right), \sigma^2 \Delta_{{g}}^2 \mathbf{I}\right) \| \mathcal{N}\left(0, \sigma^2 \Delta_{{g}}^2 \mathbf{I}\right)\right) \\
& \stackrel{(b)}{\leq} \sup _{\|{\Delta_{i}}\|_2 \leq i C} \mathcal{D}\left(\mathcal{N}\left(\Delta_i, \sigma^2 \Delta_{{g}}^2 \mathbf{I}\right) \| \mathcal{N}\left(0, \sigma^2 \Delta_{{g}}^2 \mathbf{I}\right)\right) \\
& \stackrel{(c)}=\sup _{\|\Delta_{i}\|_2 \leq i C} \frac{\alpha\|\Delta_{i}\|_2^2}{2 \Delta_{{g}}^2 \sigma^2}=\frac{\alpha i^2}{2 R_{N,L}^2\sigma^2},
\end{aligned}
\end{equation}
where $\Delta_i={{g}}_{t, i}-{{g}}_{t, i}^{\prime}$. (a) leverages the property that Rényi divergence remains unchanged under invertible transformations~\cite{34}, while (b) and (c) are derived from Theorem~\ref{theorem:L_D_NEW} and Lemma~\ref{lemma:RDP_GUASSIAN}, respectively. Based on Lemma~\ref{lemma:05} , we derive that:
\begin{equation}
\begin{aligned}
&\mathcal{D}_\alpha \left(\tilde{{g}}_{t} \| \tilde{{g}}_{t}^{\prime}\right) \leq \ln \mathbb{E}_{i \sim \beta}\left[\exp\left({\frac{\alpha i^2(\alpha-1) }{2 R_{N,L}^2 \sigma^2}}\right)\right]\\
&=\frac{1}{\alpha-1} \ln \left(\sum_{i=0}^{R_{N,L}} \beta_i \exp\left({\frac{\alpha i^2(\alpha-1) }{2 R_{N,L}^2 \sigma^2}}\right)\right)=\gamma.
\end{aligned}
\end{equation}
Here, $\beta_i$ is illustrated in Eq.~\eqref{eq:beta}. Based on the composition property of DP, after $T=n^{epoch}\cdot n^{iter}$ interations, the discriminators satisfy node-level $(\alpha, {2T}\gamma)$-RDP. Moreover, owing to the post-processing property of DP, the generators $G^+$ and $G^-$ inherit the same privacy guarantee as the discriminators. Therefore,  Algorithm~\ref{algo:ASGL} obeys node-level $(\alpha, {2T}\gamma)$-RDP, and the proof of Theorem~\ref{theorem:Privacy Analysis} is completed.

\section{Additional Details of Experiments}
\subsection{Dataset Introduction}
\label{app:data}
The detailed introduction of all datasets is as follows.
\begin{itemize}[leftmargin=5mm]
\item Bitcoin-Alpha and Bitcoin-OTC are trust networks among Bitcoin users, aimed at preventing transactions with fraudulent or high-risk users. In these networks, user relationships are represented by positive (trust) and negative (distrust) edges.
\item Slashdot is a social network derived from user interactions on a technology news site, where relationships are annotated as positive (friend) or negative (enemy) edges.
\item WikiRfA is a voting network for electing managers in Wikipedia, where edges denote positive (supporting vote) or negative (opposing vote) relationships between users.
\item  Epinions is a product review site where users can establish both trust and distrust relationships with others.
\end{itemize}

\subsection{The Distribution of Node Degrees and Path Lengths}
\label{app:NL}
\begin{figure}[!h]
\setlength{\abovecaptionskip}{2pt}
\setlength{\belowcaptionskip}{-5pt}
  \centering
  \includegraphics[width=0.9\linewidth]{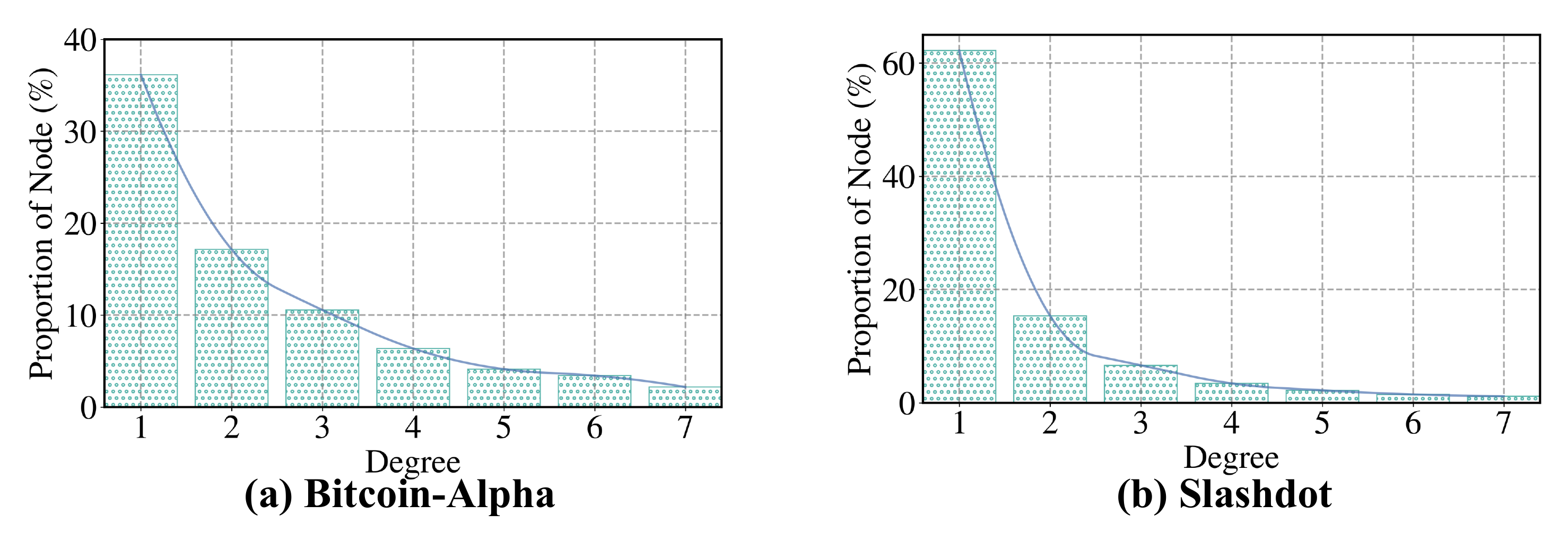}
  \caption{Distribution of node degrees.}
  \label{fig:dis_degree}
\end{figure}

\begin{figure}[!h]
\setlength{\abovecaptionskip}{2pt}
\setlength{\belowcaptionskip}{-5pt}
  \centering
  \includegraphics[width=0.9\linewidth]{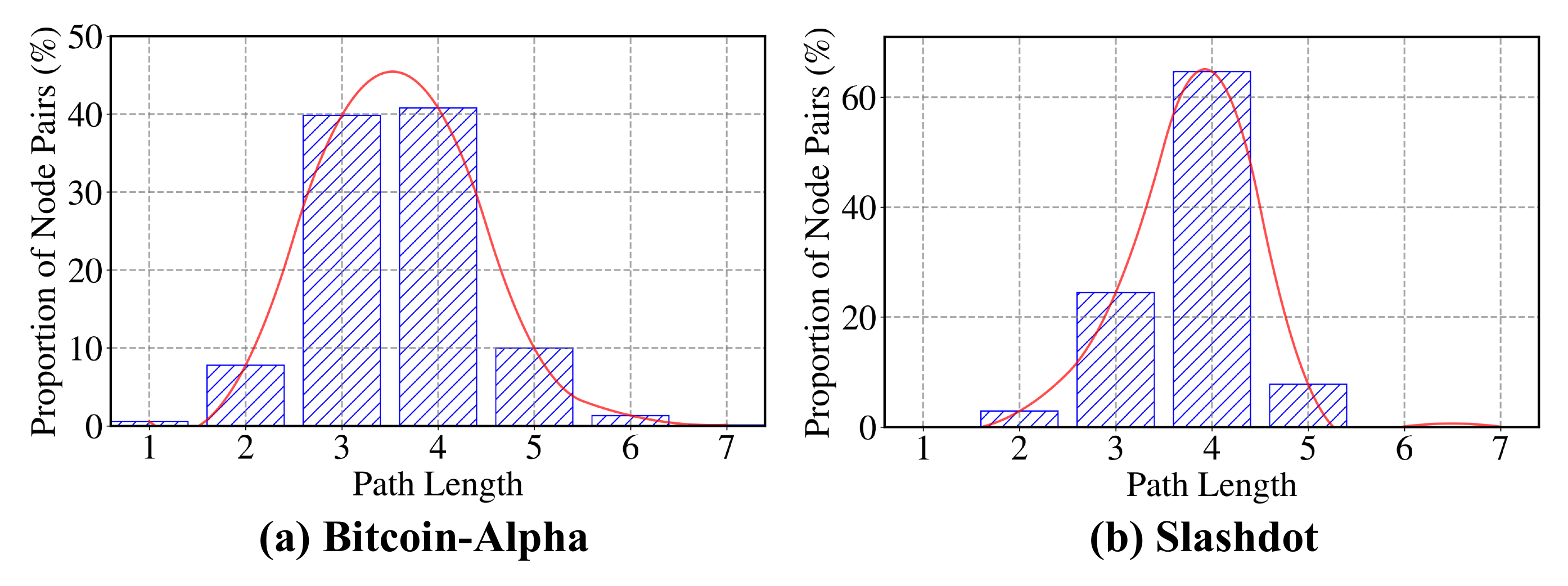}
  \caption{Distribution of path lengths.}
  \label{fig:dis_path_len}
\end{figure}

The findings for the distribution of node degrees and path lengths in the Bitcoin-Alpha and Slashdot datasets are shown in Figs.~\ref{fig:dis_degree} and~\ref{fig:dis_path_len}.

\subsection{The detailed results of Edge Sign Prediction}
\label{app:edge_sign_pre}
The average AUC results under different values of \( \epsilon \) and datasets for edge prediction tasks are detailed in Table~\ref{tab:detailed ESP}.

\begin{table}[!t]
\small
\centering
\setlength{\tabcolsep}{4pt}  
\begin{threeparttable}
  \caption{Summary of average AUC with different $\epsilon$ and datasets for edge sign prediction tasks. (\textbf{BOLD}: Best)}
  \label{tab:detailed ESP}
  \begin{tabular}{ccccccc}
    \toprule
    Dataset& Method & $\epsilon=1$ &$\epsilon=2$ &$\epsilon=3$ &$\epsilon=4$ &$\epsilon=6$\\

    \midrule
    \multirow{6}{*}{Bitcoin-OTC}
    &SDGNN &0.7655 &0.7872 &0.7913 &0.8105 &0.8571 \\
      &SiGAT &0.7011 &0.7282 &0.7869 &0.8379 &0.8706\\
      &SGCN &0.5565 &0.5740 &0.6634 &0.7516 &0.7801 \\
      &GAP &0.5763 &0.5782 &0.6486 &0.6741 &0.7411\\
    &LSNE &0.5030 &0.5405 &0.7041 &0.8239 &0.8776\\
      &ASGL &\textbf{0.8004} &\textbf{0.8462} &\textbf{0.8488} &\textbf{0.8505} &\textbf{0.8801} \\

    \midrule
    \multirow{6}{*}{Bitcoin-Alpha}
    &SDGNN &0.6761 &0.6883 &0.7098 &0.7308 &0.8476 \\
      &SiGAT &0.7033 &0.7215 &0.7303 &0.7488 &0.8207\\
      &SGCN &0.5157 &0.5450 &0.6433 &0.6930 &0.7702 \\
      &GAP &0.5664 &0.6025 &0.6367 &0.7091 &0.7320\\
    &LSNE &0.5112 &0.5361 &0.5959 &0.6524 &0.8069\\
      &ASGL &\textbf{0.7505} &\textbf{0.8075} &\textbf{0.8589} &\textbf{0.8591} &\textbf{0.8592}\\

  \midrule
    \multirow{6}{*}{WikiRfA}
    &SDGNN &0.6558 &0.7066 &0.7142 &0.7267 &0.7930 \\
      &SiGAT &0.6313 &0.6525 &0.7023 &0.7777 &0.8099\\
      &SGCN &0.5107 &0.6456 &0.6515 &0.7008 &0.7110\\
      &GAP &0.5356 &0.5506 &0.5612 &0.5717 &0.5937\\
    &LSNE &0.5086 &0.5253 &0.6119 &0.6553 &0.7832\\
      &ASGL &\textbf{0.6680} &\textbf{0.7706} &\textbf{0.7963} &\textbf{0.7986} &\textbf{0.8100} \\

    \midrule
    \multirow{6}{*}{Slashdot}
    &SDGNN &0.7547 &0.8325 &0.8697 &0.8788 &0.8862 \\
      &SiGAT &0.7061 &0.7886 &0.8392 &0.8424 &0.8527\\
      &SGCN &0.5662 &0.6151 &0.6662 &0.7181 &0.8093 \\
      &GAP &0.6121 &0.6389 &0.6879 &0.7126 &0.7471\\
    &LSNE &0.5717 &0.6144 &0.7541 &0.7753 &0.7816\\
      &ASGL &\textbf{0.7861} &\textbf{0.8539} &\textbf{0.8887} &\textbf{0.8890} &\textbf{0.8910} \\

    \midrule
    \multirow{6}{*}{Epinions}
    &SDGNN &0.6788 &0.7180 &0.7201 &0.7455 &0.8428 \\
      &SiGAT &0.6772 &0.7046 &0.7063 &0.7702 &0.8253\\
      &SGCN &0.6152 &0.6487 &0.6974 &0.7502 &0.8318 \\
      &GAP &0.5899 &0.6034 &0.6288 &0.6310 &0.6618\\
    &LSNE &0.5033 &0.6055 &0.7590 &0.8434 &0.8585\\
      &ASGL &\textbf{0.6869} &\textbf{0.8134} &\textbf{0.8513} &\textbf{0.8658} &\textbf{0.8666}\\
  \bottomrule
\end{tabular}
\end{threeparttable}
\end{table}

\subsection{The detailed results of node clustering}
\label{app:node_cluster}
The average SSI results under different values of \( \epsilon \) and datasets for node clustering tasks are detailed in Table~\ref{tab:detailed nc}.

\begin{table}[!h]  
\small
\centering
\setlength{\tabcolsep}{3pt}  
\begin{threeparttable}
\captionsetup{skip=1pt} 
  \caption{Summary of average SSI with different $\epsilon$ and datasets for node clustering tasks. (\textbf{BOLD}: Best)}
  \label{tab:detailed nc}
  \begin{tabular}{cccccccc}
    \toprule
    $\epsilon$ & Dataset &SGCN &SDGNN &SiGAT &LSNE &GAP &ASGL\\

    \midrule
    \multirow{5}{*}{1}
    &Bitcoin-Alpha &0.4819 &0.4378 &0.4877 &0.4977 &0.4988 &\textbf{0.5091}\\
    &Bitcoin-OTC &0.4505 &0.4677 &0.5025 &0.4970 &0.5008 &\textbf{0.5160}\\
    &Slashdot &0.4715 &0.5011 &0.5025 &0.5052 &0.5005 &\textbf{0.5107}\\
    &WikiRfA &0.4788 &0.4988 &0.4968 &0.4890 &0.5003 &\textbf{0.5126}\\
    &Epinions &0.5001 &0.4965 &0.5022 &0.5013 &0.6095 &\textbf{0.6106}\\

    \midrule
    \multirow{5}{*}{2}
    &Bitcoin-Alpha &0.4910 &0.4733 &0.4969 &0.4985 &0.5032 &\textbf{0.5402}\\
    &Bitcoin-OTC &0.4733 &0.4968 &0.5075 &0.4986 &0.5729 &\textbf{0.6810}\\
    &Slashdot &0.4888 &0.4864 &0.4871 &0.5134 &0.5132 &\textbf{0.5494}\\
    &WikiRfA &0.4934 &0.5054 &0.5117 &0.4996 &0.5032 &\textbf{0.5577}\\
    &Epinions &0.5068 &0.5116 &0.5086 &0.5463 &0.6263 &\textbf{0.6732}\\

    \midrule
    \multirow{5}{*}{4}
    &Bitcoin-Alpha &0.5019 &0.4948 &0.5112 &0.5049 &0.6204 &\textbf{0.6707}\\
    &Bitcoin-OTC &0.5005 &0.5325 &0.5612 &0.5465 &0.6953 &\textbf{0.7713}\\
    &Slashdot &0.5003 &0.5685 &0.5545 &0.5671 &0.5444 &\textbf{0.5994}\\
    &WikiRfA &0.5005 &0.5142 &0.5538 &0.5476 &0.5644 &\textbf{0.5977}\\
    &Epinions &0.5148 &0.5389 &0.5386 &0.6255 &0.6747 &\textbf{0.6787}\\
  \bottomrule
\end{tabular}
\end{threeparttable}
\end{table}

\begin{figure}[!t]
\setlength{\abovecaptionskip}{2pt}
\setlength{\belowcaptionskip}{-5pt}
  \centering
  \includegraphics[width=0.7\linewidth]{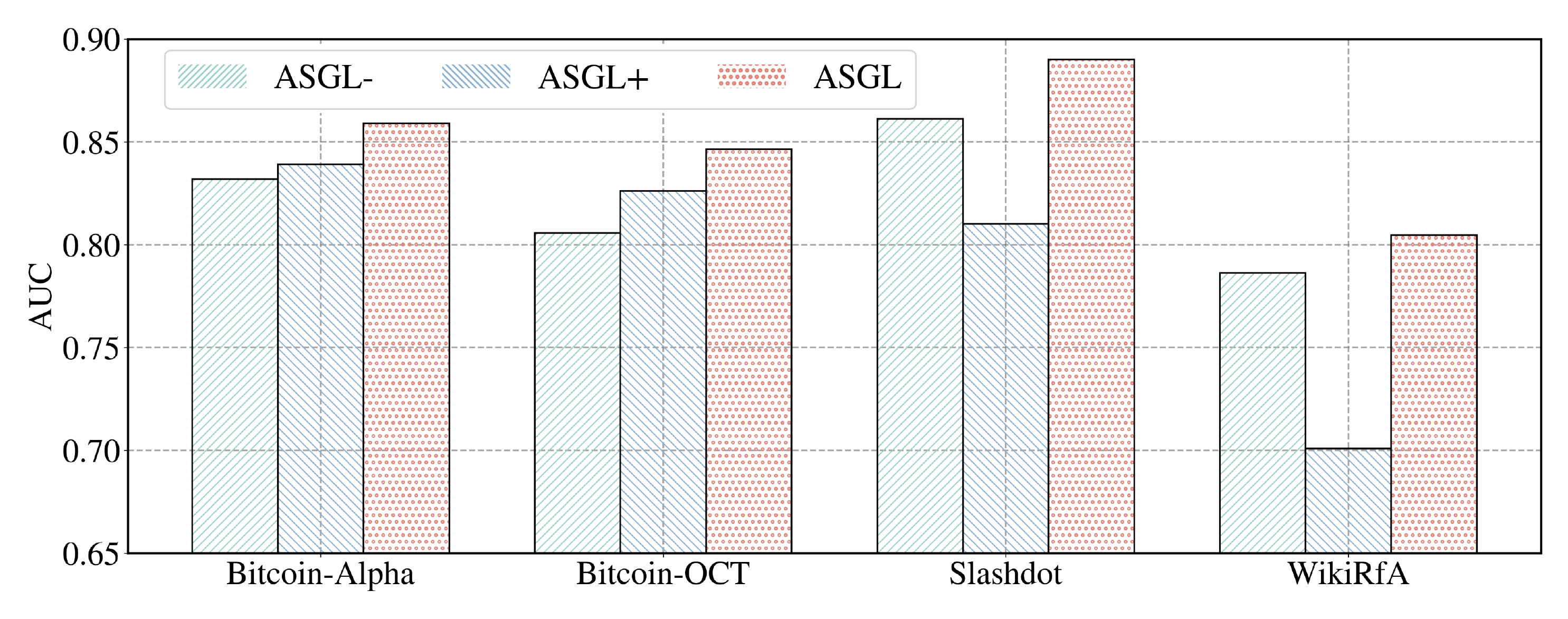}
  \caption{Comparison between ASGL, $\text{ASGL}^+$, and $\text{ASGL}^-$.}
  \label{fig:alation}
\end{figure}
\subsection{The Setup of Link Stealing Attack}
\label{sec:setup_LSA}
Motivated by~\cite{42}, we assume that the adversary has black-box access to the node embeddings produced by the target signed graph learning model, but not to its internal parameters or gradients. The adversary also possesses an auxiliary graph dataset comprising node pairs that partially overlap in distribution with the target graph. Some of these node pairs belong to the training graph (members), while others are from the test graph (non-members). For each node pair, a feature vector is constructed by concatenating their embeddings. Finally, these feature vectors, along with their corresponding member or non-member labels, are then used to train a logistic regression classifier to infer whether an edge exists between any two nodes of the target graph.
To simulate this link stealing attack, each dataset is partitioned into target training, auxiliary training, target test, and auxiliary test sets with a 5:2:2:1 ratio.

\subsection{Effectiveness of Adversarial Learning with Edge Signs.}
\label{sub:ablation}
To verify the effectiveness of adversarial learning with signed edges, we also compare our ASGL with its variants, denoted as $\text{ASGL}^+$ and $\text{ASGL}^-$. Specifically, $\text{ASGL}^+$ and $\text{ASGL}^-$ only operate on the positive graph $\mathcal{G}^+$ and the negative graph $\mathcal{G}^-$, respectively. Fig.~\ref{fig:alation} presents the average AUC scores of ASGL, $\text{ASGL}^+$, and $\text{ASGL}^-$ across all datasets. It can be observed that ASGL significantly outperforms both $\text{ASGL}^+$ and $\text{ASGL}^-$ in all cases. These results demonstrate that our privacy-preserving adversarial learning framework with edge signs is more effective in representing signed graphs compared to its variants that neglect edge sign information.

\end{document}